\pdfoutput=1

\documentclass[11pt]{article}
\usepackage{authblk}
\usepackage[utf8]{inputenc}

\usepackage[final]{acl}
\usepackage{verbatim}
\usepackage{times}
\usepackage{latexsym}
\usepackage{booktabs}
\usepackage{tabulary}

\usepackage[draft]{minted}

\usepackage{hyperref,xcolor}
\usepackage{supertabular,booktabs}

\usepackage{hyperref}

\usepackage[T1]{fontenc}

\usepackage[utf8]{inputenc}

\usepackage{microtype}
\usepackage{longtable}

\newcommand\N{480}

\title{tasksource: A Dataset Harmonization Framework \\
for Streamlined NLP Multi-Task Learning and Evaluation}

\author[1]{Damien Sileo}

\affil[1]{Univ. Lille, Inria, CNRS, Centrale Lille, UMR 9189 - CRIStAL, F-59000 Lille, France}
\affil[ ]{\url{damien.sileo@inria.fr}}

\begin{document}

\maketitle
\begin{abstract}
The HuggingFace Datasets Hub hosts thousands of datasets, offering exciting opportunities for language model training and evaluation. However, datasets for a specific task type often have different schemas, making harmonization challenging\footnote{\url{https://xkcd.com/927/}}. Multi-task training or evaluation necessitates manual work to fit data into task templates. Several initiatives independently tackle this issue by releasing harmonized datasets or providing harmonization codes to preprocess datasets into a consistent format. We identify patterns across previous preprocessing efforts, such as column name mapping and extracting specific sub-fields from structured data in a column. We then propose a structured annotation framework that ensures our annotations are fully exposed and not hidden within unstructured code. We release a dataset annotation framework and dataset annotations for more than 500 English tasks\footnote{\url{https://github.com/sileod/tasksource}}. These annotations include metadata, such as the names of columns to be used as input or labels for all datasets, which can save time for future dataset preprocessing, regardless of whether our framework is utilized. We fine-tune a multi-task text encoder on all tasksource tasks, outperforming every publicly available text encoder of comparable size in an external evaluation\footnote{\href{https://hf.co/sileod/deberta-v3-base-tasksource-nli}{\texttt{hf.co/sileod/deberta-v3-base-tasksource-nli}}}.

\end{abstract}

\section{Introduction}

Datasets are a key ingredient in modern artificial natural language processing (NLP). 

Language understanding models trained on unannotated corpora need to be evaluated, and individual datasets or benchmarks with multiple datasets provide an objective measure of targeted model capabilities. Supervised fine-tuning on annotated datasets also leads to better evaluations, and multi-task learning (MTL) \citep{Caruana1993MultitaskLA} or extreme MTL \citep{aribandi2022ext}, i.e. MTL with many tasks, improves robustness.

The HuggingFace Datasets \citep{2020HuggingFace-datasets} Hub hosts thousands of datasets. However, running evaluations or MTL on many datasets requires manual work because of a lack of standardization. Fine-tuning a model on multiple datasets requires alignment of datasets formats, even with a single task type (e.g. natural language inference). Because of that, various initiatives assemble datasets or preprocessing code to ease multi-task learning or benchmarking. However, they either distribute prepossessed copies of the datasets or preprocessing code associated with each dataset. Section \ref{sec:relatedwork} enumerates previous works enacting these two approaches.

The previous preprocessing codes implicitly use some metadata, such as mappings between column names and fields of a task, but extracting it is quite difficult. Code is not disentangled from metadata.
We propose a very concise dataset annotation format by relying on patterns reoccurring across several preprocessings. Most annotations fit in a single line, e.g:
\begin{minted}{python}
scitail = Classification(
  'sentence1',
  'sentence2',
  'gold_label')
\end{minted} 
The SciTail \citep{scitail} dataset on HF-Hub, noted \texttt{scitail\_ds}\footnote{\url{https://hf.co/datasets/scitail}} \citep{scitail} can be standardized by calling the \texttt{tasksource.scitail} function, and associated metadata can be retrieved with \mintinline{python}|tasksource.scitail.dict()|.

We annotate \N\ tasks, focusing on discriminative tasks to complement previous work better. We train a \texttt{deberta-base} text encoder on all of them simultaneously (Section \ref{sec:model}) leading to unprecedented model performance (Section \ref{sec:results}).

\section{Related work \label{sec:relatedwork}}
Various works harmonize existing datasets by either sharing preprocessed copies or preprocessings. Tasksource is a collection of preprocessings, and it is the largest for tasks excluding text generation tasks.
Text generations tasks have a relatively simple format (optional input text, and output text), and previous work such as  PromptSource \citep{bach2022promptsource} and SuperNatural Instructions\cite{supernaturalinstructions} did not provide structured annotations, as defined in section \ref{sec:datasetparsing}, but these can still be combined with acceptable efforts. 

\paragraph{Preprocessed copies}
BIG-Bench \citep{bigbench}, BigBio \citep{fries2022bigbio}, Natural and SuperNatural Instructions \citep{naturalinstructions,supernaturalinstructions}, PragmEval \citep{sileo-etal-2022-pragmatics}, UnifiedQA \citep{2020unifiedqa}, TweetEval \citep{barbieri2020tweeteval}, DiscoEval \citep{mchen-discoeval-19}, Silicone \citep{chapuis-etal-2020-hierarchical}, LexGLUE \citep{chalkidis-etal-2022-lexglue}, SetFit 
 \citep{tunstall2022efficient} distribute preprocessed copies of the original data with standardized format.

\paragraph{Collections of preprocessings}
SentEval \citep{conneau-etal-2017-supervised},  Jiant \citep{pruksachatkun-etal-2020-jiant}, BLUE \cite{peng2019transfer}, MetaEval \citep{sileo-moens-2022-analysis}, CrossFit \citep{ye-etal-2021-crossfit}, PromptSource \citep{bach2022promptsource} distribute the code required to jointly use some datasets with initially distinct structures. ExMix \citep{aribandi2022ext} is not released to our knowledge. The Muppet \citep{aghajanyan-etal-2021-muppet} authors did not release their preprocessing either.

Our work also pertains to extreme MTL \citep{aribandi2022ext,aghajanyan-etal-2021-muppet} and dataset count scaling.

\section{Structured dataset annotation \label{sec:datasetparsing}}

We define \textit{dataset parsing} as the mapping of a dataset into a task template. 

A task template is a type of task, like paraphrase detection, associated with a predetermined set of fields. For example, Paraphrase detection can be mapped to a task template \textsc{ParaphraseDetection(Sentence1, Sentence2, Label)}

A dataset is a set of examples with named and typed columns. \texttt{quora} is an example of a dataset hosted on the HuggingFace Datasets Hub \citep{2020HuggingFace-datasets}, illustrated in Table \ref{tab:quora}.

\begin{table}[h]
\begin{tabulary}{.85\linewidth}{LCL}
\toprule
\textbf{questions (sequence)} &
\textbf{is\_duplicate (bool)} \\ \midrule
\{ "id": {[} 1, 2 {]}, "text": {[} "What is the step by step guide to invest in share market in india?", "What is the step by step guide to invest in share market?" {]} \} &
  false \\
...& \\ \bottomrule
\end{tabulary}
\caption{One row of the Quora dataset, as hosted on the HuggingFace Datasets Hub. \label{tab:quora}}
\end{table}

A dataset parser for a specific dataset is a function that maps the whole dataset, or examples, to a task format, which can be \textsc{ParaphraseDetection} here.

As seen above, some benchmarks distribute harmonized datasets. This approach can save computations but can waste storage space, and make it harder to track all the design decisions that were applied to the original dataset. Users can also implement parsers themselves, or rely on external libraries to process examples on a restricted set of tasks. The previous preprocessings codes do not disentangle data and logic, and cannot be seen as semantic dataset annotations. This complicates the combinations of different preprocessing. Previous preprocessings also contain repetitive boilerplate code\footnote{i.e. \url{https://github.com/INK-USC/CrossFit/blob/master/tasks/aqua_rat.py}}.

We decompose  dataset parsing logic from annotations with  based on two observations:

(1) The fields of task type (e.g. \textsc{Sentence1},\textsc{Sentence2} for \textsc{ParaphraseDetection} can often be independently mapped to functions of dataset examples. Therefore, we can annotate a dataset with a task type, then annotate each field of a task type with a function that extracts the desired information from a dataset example.

(2) The field mapping functions are often selecting a column from examples: in that case, they can be annotated with the name of the relevant columns. Sometimes, as in the Quora dataset in Table \ref{tab:quora}, they are selecting a path from a nested structure: in that case, they can be annotated with a path. Fields can also be mapped to a constant -- some multiple-choice question-answering datasets always use the first choice as the correct choice and have an implicit constant label equal to 0. A field can also be mapped to a concatenation of the text of different columns, which can also be annotated with parameters.


\section{Tasksource dataset annotation format}

Once a dataset is annotated with a task type, each field of the task type has to be annotated with a function that takes an example from the dataset and returns the intended part of the example.
For brevity, we can annotate a field with a string \texttt{s} to denote the function \mintinline{python}|lambda x:x[s]|

The tasksource backend handles the annotations and turns them into harmonizing preprocessing. We consider 3 general task types:

\textsc{Classification(text1, text2, labels)} where \textsc{LABELS} has to be a function that takes an example and returns a class index. It can also return a float for regression tasks, or a fixed-size list for multi-label classification. 
\textsc{text1} takes a dataset example as input and returns the text extracted from the example.
\textsc{text2} is optional and is here to leverage the fact that most text encoders process text pairs with special care.

\textsc{MultipleChoice(prompt, 
choices, labels)}:
\textsc{choices} has to be a function that returns a list of text choices (the number of choices can differ across examples) extracted from an example. For concision, it can also be a list of column names to denote a list of textual choices already available in the example.  \textsc{labels} has to return the index of the correct choice (most tasks have only one correct answer).

\textsc{TokenClassification(tokens, labels)} where \textsc{tokens} takes an example as input and returns to a list of already split tokens,  \textsc{labels} return a list of labels aligned to the tokens ($i^{th}$ label annotates the $i^{th}$ token).


We also provide 3 structured function factories to cover additional use cases while exposing their behavior with parameters.

 \paragraph{\texttt{get}} enables to access nested objects. \mintinline{python}|get.questions.text[0]| is equivalent to \mintinline{python}|lambda x:x['questions']['text'][0]|

\paragraph{\texttt{constant}} provides constant functions. \texttt{constant(x)} is equivalent to \mintinline{python}|lambda *_:x|.

\paragraph{\texttt{cat}} concatenates multiple columns that contain strings. \mintinline{python}|cat(col1 col2)| is equivalent to \mintinline{python}|lambda x:x[col1]+x[col2]|.

\mbox{} 

An annotation to parse the Quora dataset in Table \ref{tab:quora} can then be written as follows:
\begin{minted}{python}
quora = Classification(
  text1=get.questions.text[0], 
  text2=get.questions.text[1],
  labels='is_duplicate')
\end{minted}

For completeness, we also allow optional \texttt{preprocess} and \texttt{postprocess} arguments to a task type. They should be functions that take the full dataset as input and return a dataset. We found this feature to be necessary in a few cases where datasets had unusable labels (e.g. negative label indexes) that caused errors, or to edit the metadata of a dataset, like the name of the labels when it needs to be changed.

\section{Tasksource annotations}
We select English datasets available on the HuggingFace Datasets Hub. We only consider discriminative tasks (Classification, Multiple-choice, Token Classification). We crawled all the tasks tagged with the English Language, and the Text-Classification task type\footnote{\url{https://hf.co/datasets?language=language:en&task_categories=task_categories:text-classification&sort=downloads}}
or Multiple Choice tag\footnote{\url{https://hf.co/datasets?task_categories=task_categories:multiple-choice&sort=downloads}}, as of January 2023.

As many tags are missing, to increase the coverage, we crawled the 1000 most popular datasets and used heuristics to identify discriminative tasks with labels with their fields names. We then ran a \texttt{fasttext} \citep{joulin2016fasttext} \texttt{langid} classifier to filter out untagged datasets with non-English text.

We only annotate datasets that do not require the user to manually download data or sign an agreement. We exclude datasets that require a particular library, with the exception of BIG-bench.
We also exclude tasks where high accuracy is not desirable, such as bias probing tasks \citep{nangia-etal-2020-crows} where accuracy measures bias, and tasks with input length that mostly exceeds 256 tokens.

 We  manually deduplicate the datasets which can be available individually or in benchmarks.

We also annotate the mapping between split names and train/validation/test splits. When the test splits are obfuscated (labels unavailable), we split the validation set and use half of it as a test set. Our goal is to reduce friction and individually submitting model test predictions to data owners can take a lot of time.
When no split is available, we do a 80/10/10\% split. We use a fixed 0 random seed. to help reproducibility.

Label handling was one of the pain points of the testing of the preprocessing functions. The tasksource backend preprocesses text labels to map them to integers. 

The Table in Appendix \ref{sec:appendix} enumerates all datasets annotated in the current version of tasksource\footnote{Annotations: \url{https://github.com/sileod/tasksource/blob/main/src/tasksource/tasks.py}}

\section{Pretraining a model on tasksource \label{sec:model}}
To demonstrate the potential of tasksource, we fine-tune a single \texttt{deberta-base-v3} \citep{he2021debertav3}\footnote{This is the best-performing unsupervisedly pretrained text encoder of this size according to the GLUE Benchmark \citep{wang2019glue}.} text encoder on all tasksource tasks.

Following BERT \citep{devlin-etal-2019-bert} standard setup, for token-classification tasks, we use a softmax classifier on top of the last layer encoded tokens to predict the token classes. For classification tasks and multiple-choice tasks, we use a classifier on top of the \texttt{[CLS]} sentinel token last layer.

We assign each task a different classification layer, but we tie the label weights (not biases) to each other if they are all identical. 

We oversample datasets by a factor of 2 if they have less than $64k$ examples then cap dataset size to $64k$ examples to foster dataset diversity. We randomly sample a task for each batch with a frequency proportional to the capped training dataset size and we add a learnable task-specific sentinel token to the shared sentinel token. We drop the task-specific token $10\%$ of the time to teach the model to also work without these task embeddings, to reduce mismatch when using our model with the vanilla DeBERTa architecture.  We also noticed that this tended to improve general accuracy, since this forces cooperation across tasks.

We limit the number of choices to 4 for multiple-choice tasks, to limit redundant computations, as some datasets have more than 100 choices.

We use a learning rate of $3.10^{-5}$, a sequence length of 256, and a batch size of 24, with 16 accumulation steps to stabilize the multi-task optimization \citep{yu2020gradient}. We did not perform hyperparameter optimization.

We used the \texttt{tasknet}\footnote{Tasknet \citep{sileod22-tasknet} is 
 interface Huggingface Datasets with Huggingface Trainer.} library and a single RTX-6000 24GB GPU for 7 days (20k steps).
Using tasksource with tasknet enables concise multi-task training\footnote{\url{https://colab.research.google.com/drive/1iB4Oxl9_B5W3ZDzXoWJN-olUbqLBxgQS?usp=sharing}}.
\section{Results \label{sec:results}}

As of January 2023, an early version of our model ranks first among 3574 base-sized\footnote{This corresponds to 86M encoder parameters excluding embeddings.} model on  the \textit{Model Recycling} \citep{choshen2022start} external evaluation\footnote{\url{https://ibm.github.io/model-recycling}}
This evaluation comprises 36 representative English NLP tasks (Consisting of sentiment, NLI, Twitter, topic classification, and other general classification tasks), over 5 random seeds. These results are competitive with \texttt{deberta-large} models on GLUE. We did not observe any sign of over-fitting  yet which suggests that the network might still be undertrained.

\section{Conclusion}

We described a semantic, structured, concise, expressive dataset preprocessing annotation framework, which is associated with a parser and annotations, that can greatly facilitate new experiments for multi-task learning and improve reproducibility. We only scratched the surface of the potential of this generated task collection due to computational limitations. For future work, we plan to use tasksource to fully automate dataset parsing on new datasets with machine learning techniques to learn the parsing process. We also plan to work on a multilingual extension of tasksource annotations.

\bibliography{all,all_addendum}

\begin{thebibliography}{203}
\expandafter\ifx\csname natexlab\endcsname\relax\def\natexlab#1{#1}\fi

\bibitem[{ACL(2017)}]{ACL}
 2017.
\newblock Program induction by rationale generation: Learning to solve and
  explain algebraic word problems.

\bibitem[{Zur(2017)}]{ZurichOpenRepositoryandArchive:dataset}
 2017.
\newblock Software applications user reviews.

\bibitem[{ZRo(2018)}]{ZRoshanSharma:dataset}
 2018.
\newblock Sentimental analysis of tweets for detecting hate/racist speeches.

\bibitem[{ai2(2019)}]{ai2:winogrande}
 2019.
\newblock Winogrande: An adversarial winograd schema challenge at scale.

\bibitem[{Aghahadi and Talebpour(2022)}]{aghahadi2022avicenna}
Zeinab Aghahadi and Alireza Talebpour. 2022.
\newblock Avicenna: a challenge dataset for natural language generation toward
  commonsense syllogistic reasoning.
\newblock \emph{Journal of Applied Non-Classical Logics}, pages 1--17.

\bibitem[{Aghajanyan et~al.(2021)Aghajanyan, Gupta, Shrivastava, Chen,
  Zettlemoyer, and Gupta}]{aghajanyan-etal-2021-muppet}
Armen Aghajanyan, Anchit Gupta, Akshat Shrivastava, Xilun Chen, Luke
  Zettlemoyer, and Sonal Gupta. 2021.
\newblock \href {https://doi.org/10.18653/v1/2021.emnlp-main.468} {Muppet:
  Massive multi-task representations with pre-finetuning}.
\newblock In \emph{Proceedings of the 2021 Conference on Empirical Methods in
  Natural Language Processing}, pages 5799--5811, Online and Punta Cana,
  Dominican Republic. Association for Computational Linguistics.

\bibitem[{Almeida et~al.(2011)Almeida, Hidalgo, and
  Yamakami}]{Almeida2011SpamFiltering}
Tiago~A. Almeida, Jose Maria~Gomez Hidalgo, and Akebo Yamakami. 2011.
\newblock Contributions to the study of sms spam filtering: New collection and
  results.
\newblock In \emph{Proceedings of the 2011 ACM Symposium on Document
  Engineering (DOCENG'11)}.

\bibitem[{Aribandi et~al.(2022)Aribandi, Tay, Schuster, Rao, Zheng, Mehta,
  Zhuang, Tran, Bahri, Ni, Gupta, Hui, Ruder, and Metzler}]{aribandi2022ext}
Vamsi Aribandi, Yi~Tay, Tal Schuster, Jinfeng Rao, Huaixiu~Steven Zheng,
  Sanket~Vaibhav Mehta, Honglei Zhuang, Vinh~Q. Tran, Dara Bahri, Jianmo Ni,
  Jai Gupta, Kai Hui, Sebastian Ruder, and Donald Metzler. 2022.
\newblock \href {https://openreview.net/forum?id=Vzh1BFUCiIX} {Ext5: Towards
  extreme multi-task scaling for transfer learning}.
\newblock In \emph{International Conference on Learning Representations}.

\bibitem[{Aroca-Ouellette et~al.(2021)Aroca-Ouellette, Paik, Roncone, and
  Kann}]{aroca-ouellette-etal-2021-prost}
Stephane Aroca-Ouellette, Cory Paik, Alessandro Roncone, and Katharina Kann.
  2021.
\newblock \href {https://aclanthology.org/2021.findings-acl.404} {{PROST}:
  {P}hysical reasoning about objects through space and time}.
\newblock In \emph{Findings of the Association for Computational Linguistics:
  ACL-IJCNLP 2021}, pages 4597--4608, Online. Association for Computational
  Linguistics.

\bibitem[{Asher et~al.(2016)Asher, Hunter, Morey, Farah, and
  Afantenos}]{asher-etal-2016-discourse}
Nicholas Asher, Julie Hunter, Mathieu Morey, Benamara Farah, and Stergos
  Afantenos. 2016.
\newblock \href {https://www.aclweb.org/anthology/L16-1432} {Discourse
  structure and dialogue acts in multiparty dialogue: the {STAC} corpus}.
\newblock In \emph{Proceedings of the Tenth International Conference on
  Language Resources and Evaluation ({LREC}'16)}, pages 2721--2727, Portoro{
  {z}}, Slovenia. European Language Resources Association (ELRA).

\bibitem[{Bach et~al.(2022)Bach, Sanh, Yong, Webson, Raffel, Nayak, Sharma,
  Kim, Bari, Fevry, Alyafeai, Dey, Santilli, Sun, Ben-David, Xu, Chhablani,
  Wang, Fries, Al-shaibani, Sharma, Thakker, Almubarak, Tang, Tang, Jiang, and
  Rush}]{bach2022promptsource}
Stephen~H. Bach, Victor Sanh, Zheng-Xin Yong, Albert Webson, Colin Raffel,
  Nihal~V. Nayak, Abheesht Sharma, Taewoon Kim, M~Saiful Bari, Thibault Fevry,
  Zaid Alyafeai, Manan Dey, Andrea Santilli, Zhiqing Sun, Srulik Ben-David,
  Canwen Xu, Gunjan Chhablani, Han Wang, Jason~Alan Fries, Maged~S.
  Al-shaibani, Shanya Sharma, Urmish Thakker, Khalid Almubarak, Xiangru Tang,
  Xiangru Tang, Mike Tian-Jian Jiang, and Alexander~M. Rush. 2022.
\newblock \href {http://arxiv.org/abs/2202.01279} {Promptsource: An integrated
  development environment and repository for natural language prompts}.

\bibitem[{Bao et~al.(2022)Bao, Peng, Hartill, Tan, Deng, Witbrock, and
  Liu}]{bao2022multi}
Qiming Bao, Alex~Yuxuan Peng, Tim Hartill, Neset Tan, Zhenyun Deng, Michael
  Witbrock, and Jiamou Liu. 2022.
\newblock Multi-step deductive reasoning over natural language: An empirical
  study on out-of-distribution generalisation.
\newblock The 2nd International Joint Conference on Learning and Reasoning and
  16th International Workshop on Neural-Symbolic Learning and Reasoning
  (IJCLR-NeSy 2022).

\bibitem[{Barbieri et~al.(2020)Barbieri, Camacho-Collados, Espinosa-Anke, and
  Neves}]{barbieri2020tweeteval}
Francesco Barbieri, Jose Camacho-Collados, Luis Espinosa-Anke, and Leonardo
  Neves. 2020.
\newblock {TweetEval:Unified Benchmark and Comparative Evaluation for Tweet
  Classification}.
\newblock In \emph{Proceedings of Findings of EMNLP}.

\bibitem[{Ben~Zhou and Roth(2019)}]{ZKNR19}
Qiang~Ning Ben~Zhou, Daniel~Khashabi and Dan Roth. 2019.
\newblock “going on a vacation” takes longer than “going for a walk”: A
  study of temporal commonsense understanding.
\newblock In \emph{EMNLP}.

\bibitem[{Berzak et~al.(2020)Berzak, Malmaud, and Levy}]{starc2020}
Yevgeni Berzak, Jonathan Malmaud, and Roger Levy. 2020.
\newblock Starc: Structured annotations for reading comprehension.
\newblock In \emph{ACL}. Association for Computational Linguistics.

\bibitem[{Bhakthavatsalam et~al.(2020)Bhakthavatsalam, Richardson, Tandon, and
  Clark}]{bhakthavatsalam2020dogs}
Sumithra Bhakthavatsalam, Kyle Richardson, Niket Tandon, and Peter Clark. 2020.
\newblock \href {http://arxiv.org/abs/2006.07510} {Do dogs have whiskers? a new
  knowledge base of haspart relations}.

\bibitem[{Bisk et~al.(2020)Bisk, Zellers, Bras, Gao, and Choi}]{Bisk2020}
Yonatan Bisk, Rowan Zellers, Ronan~Le Bras, Jianfeng Gao, and Yejin Choi. 2020.
\newblock Piqa: Reasoning about physical commonsense in natural language.
\newblock In \emph{Thirty-Fourth AAAI Conference on Artificial Intelligence}.

\bibitem[{Borkan et~al.(2019)Borkan, Dixon, Sorensen, Thain, and
  Vasserman}]{DBLP:journals/corr/abs-1903-04561}
Daniel Borkan, Lucas Dixon, Jeffrey Sorensen, Nithum Thain, and Lucy Vasserman.
  2019.
\newblock \href {http://arxiv.org/abs/1903.04561} {Nuanced metrics for
  measuring unintended bias with real data for text classification}.
\newblock \emph{CoRR}, abs/1903.04561.

\bibitem[{Bowman et~al.(2015)Bowman, Angeli, Potts, and
  Manning}]{snli:emnlp2015}
Samuel~R. Bowman, Gabor Angeli, Christopher Potts, and Christopher~D. Manning.
  2015.
\newblock A large annotated corpus for learning natural language inference.
\newblock In \emph{Proceedings of the 2015 Conference on Empirical Methods in
  Natural Language Processing (EMNLP)}. Association for Computational
  Linguistics.

\bibitem[{Buechel and Hahn(2017)}]{buechel-hahn-2017-emobank}
Sven Buechel and Udo Hahn. 2017.
\newblock \href {https://www.aclweb.org/anthology/E17-2092} {{E}mo{B}ank:
  Studying the impact of annotation perspective and representation format on
  dimensional emotion analysis}.
\newblock In \emph{Proceedings of the 15th Conference of the {E}uropean Chapter
  of the Association for Computational Linguistics: Volume 2, Short Papers},
  pages 578--585, Valencia, Spain. Association for Computational Linguistics.

\bibitem[{Busso et~al.(2008)Busso, Bulut, Lee, Kazemzadeh, Mower, Kim, Chang,
  Lee, and Narayanan}]{busso2008iemocap}
Carlos Busso, Murtaza Bulut, Chi-Chun Lee, Abe Kazemzadeh, Emily Mower, Samuel
  Kim, Jeannette~N Chang, Sungbok Lee, and Shrikanth~S Narayanan. 2008.
\newblock Iemocap: Interactive emotional dyadic motion capture database.
\newblock \emph{Language resources and evaluation}, 42(4):335.

\bibitem[{Cao and Wang(2021)}]{cao-wang-2021-controllable}
Shuyang Cao and Lu~Wang. 2021.
\newblock \href {https://doi.org/10.18653/v1/2021.acl-long.502} {Controllable
  open-ended question generation with a new question type ontology}.
\newblock In \emph{Proceedings of the 59th Annual Meeting of the Association
  for Computational Linguistics and the 11th International Joint Conference on
  Natural Language Processing (Volume 1: Long Papers)}, pages 6424--6439,
  Online. Association for Computational Linguistics.

\bibitem[{Carlile et~al.(2018)Carlile, Gurrapadi, Ke, and
  Ng}]{Persuasion2018Ng}
Winston Carlile, Nishant Gurrapadi, Zixuan Ke, and Vincent Ng. 2018.
\newblock \href {https://doi.org/10.18653/v1/P18-1058} {Give me more feedback:
  Annotating argument persuasiveness and related attributes in student essays}.
\newblock In \emph{Proceedings of the 56th Annual Meeting of the Association
  for Computational Linguistics (Volume 1: Long Papers)}, pages 621--631,
  Melbourne, Australia. Association for Computational Linguistics.

\bibitem[{Caruana(1993)}]{Caruana1993MultitaskLA}
Rich Caruana. 1993.
\newblock Multitask learning: A knowledge-based source of inductive bias.
\newblock In \emph{International Conference on Machine Learning}.

\bibitem[{Casanueva et~al.(2020)Casanueva, Temcinas, Gerz, Henderson, and
  Vulic}]{Casanueva2020}
Inigo Casanueva, Tadas Temcinas, Daniela Gerz, Matthew Henderson, and Ivan
  Vulic. 2020.
\newblock \href {https://arxiv.org/abs/2003.04807} {Efficient intent detection
  with dual sentence encoders}.
\newblock In \emph{Proceedings of the 2nd Workshop on NLP for ConvAI - ACL
  2020}.
\newblock Data available at
  https://github.com/PolyAI-LDN/task-specific-datasets.

\bibitem[{Cer et~al.(2017)Cer, Diab, Agirre, Lopez-Gazpio, and
  Specia}]{cer-etal-2017-semeval}
Daniel Cer, Mona Diab, Eneko Agirre, Inigo Lopez-Gazpio, and Lucia Specia.
  2017.
\newblock \href {https://doi.org/10.18653/v1/S17-2001} {{S}em{E}val-2017 task
  1: Semantic textual similarity multilingual and crosslingual focused
  evaluation}.
\newblock In \emph{Proceedings of the 11th International Workshop on Semantic
  Evaluation ({S}em{E}val-2017)}, pages 1--14, Vancouver, Canada. Association
  for Computational Linguistics.

\bibitem[{Chakravarthi(2020)}]{chakravarthi-2020-hopeedi}
Bharathi~Raja Chakravarthi. 2020.
\newblock \href {https://www.aclweb.org/anthology/2020.peoples-1.5}
  {{H}ope{EDI}: A multilingual hope speech detection dataset for equality,
  diversity and inclusion}.
\newblock In \emph{Proceedings of the Third Workshop on Computational Modeling
  of People's Opinions, Personality and Emotion's in Social Media}, pages
  41--53, Barcelona, Spain (Online). Association for Computational Linguistics.

\bibitem[{Chalkidis et~al.(2021)Chalkidis, Fergadiotis, and
  Androutsopoulos}]{chalkidis-etal-2021-multieurlex}
Ilias Chalkidis, Manos Fergadiotis, and Ion Androutsopoulos. 2021.
\newblock Multieurlex -- a multi-lingual and multi-label legal document
  classification dataset for zero-shot cross-lingual transfer.
\newblock In \emph{Proceedings of the 2021 Conference on Empirical Methods in
  Natural Language Processing}.

\bibitem[{Chalkidis et~al.(2022)Chalkidis, Jana, Hartung, Bommarito,
  Androutsopoulos, Katz, and Aletras}]{chalkidis-etal-2022-lexglue}
Ilias Chalkidis, Abhik Jana, Dirk Hartung, Michael Bommarito, Ion
  Androutsopoulos, Daniel Katz, and Nikolaos Aletras. 2022.
\newblock \href {https://doi.org/10.18653/v1/2022.acl-long.297} {{L}ex{GLUE}: A
  benchmark dataset for legal language understanding in {E}nglish}.
\newblock In \emph{Proceedings of the 60th Annual Meeting of the Association
  for Computational Linguistics (Volume 1: Long Papers)}, pages 4310--4330,
  Dublin, Ireland. Association for Computational Linguistics.

\bibitem[{Chandra et~al.(2020)Chandra, Ronan, Chaitanya, Keisuke, Ari, Hannah,
  Doug, Scott, and Yejin}]{anli}
Bhagavatula Chandra, Le~Bras Ronan, Malaviya Chaitanya, Sakaguchi Keisuke,
  Holtzman Ari, Rashkin Hannah, Downey Doug, Wen-tau~Yih Scott, and Choi Yejin.
  2020.
\newblock Abductive commonsense reasoning.

\bibitem[{Chapuis et~al.(2020)Chapuis, Colombo, Manica, Labeau, and
  Clavel}]{chapuis-etal-2020-hierarchical}
Emile Chapuis, Pierre Colombo, Matteo Manica, Matthieu Labeau, and Chloe
  Clavel. 2020.
\newblock \href {https://doi.org/10.18653/v1/2020.findings-emnlp.239}
  {Hierarchical pre-training for sequence labelling in spoken dialog}.
\newblock In \emph{Findings of the Association for Computational Linguistics:
  EMNLP 2020}, pages 2636--2648, Online. Association for Computational
  Linguistics.

\bibitem[{Chatterjee et~al.(2019)Chatterjee, Narahari, Joshi, and
  Agrawal}]{chatterjee-etal-2019-semeval}
Ankush Chatterjee, Kedhar~Nath Narahari, Meghana Joshi, and Puneet Agrawal.
  2019.
\newblock \href {https://doi.org/10.18653/v1/S19-2005} {Semeval-2019 task 3:
  Emocontext contextual emotion detection in text}.
\newblock In \emph{Proceedings of the 13th International Workshop on Semantic
  Evaluation}, pages 39--48, Minneapolis, Minnesota, USA. Association for
  Computational Linguistics.

\bibitem[{Chen et~al.(2019{\natexlab{a}})Chen, DArcy, Liu, Fernandez, and
  Downey}]{chen2019codah}
Michael Chen, Mike DArcy, Alisa Liu, Jared Fernandez, and Doug Downey.
  2019{\natexlab{a}}.
\newblock Codah: An adversarially-authored question answering dataset for
  common sense.
\newblock In \emph{Proceedings of the 3rd Workshop on Evaluating Vector Space
  Representations for NLP}, pages 63--69.

\bibitem[{Chen et~al.(2019{\natexlab{b}})Chen, Chu, and
  Gimpel}]{mchen-discoeval-19}
Mingda Chen, Zewei Chu, and Kevin Gimpel. 2019{\natexlab{b}}.
\newblock Evaluation benchmarks and learning criteria for discourse-aware
  sentence representations.
\newblock In \emph{Proc. of {EMNLP}}.

\bibitem[{Chen et~al.(2018)Chen, Hsu, Kuo, Ku et~al.}]{chen2018emotionlines}
Sheng-Yeh Chen, Chao-Chun Hsu, Chuan-Chun Kuo, Lun-Wei Ku, et~al. 2018.
\newblock Emotionlines: An emotion corpus of multi-party conversations.
\newblock \emph{arXiv preprint arXiv:1802.08379}.

\bibitem[{Choshen et~al.(2022)Choshen, Venezian, Don-Yehia, Slonim, and
  Katz}]{choshen2022start}
Leshem Choshen, Elad Venezian, Shachar Don-Yehia, Noam Slonim, and Yoav Katz.
  2022.
\newblock Where to start? analyzing the potential value of intermediate models.
\newblock \emph{arXiv preprint arXiv:2211.00107}.

\bibitem[{Clark et~al.(2019)Clark, Lee, Chang, Kwiatkowski, Collins, and
  Toutanova}]{clark2019boolq}
Christopher Clark, Kenton Lee, Ming-Wei Chang, Tom Kwiatkowski, Michael
  Collins, and Kristina Toutanova. 2019.
\newblock Boolq: Exploring the surprising difficulty of natural yes/no
  questions.
\newblock In \emph{NAACL}.

\bibitem[{Clark et~al.(2018)Clark, Cowhey, Etzioni, Khot, Sabharwal, Schoenick,
  and Tafjord}]{allenai:arc}
Peter Clark, Isaac Cowhey, Oren Etzioni, Tushar Khot, Ashish Sabharwal, Carissa
  Schoenick, and Oyvind Tafjord. 2018.
\newblock Think you have solved question answering? try arc, the ai2 reasoning
  challenge.
\newblock \emph{arXiv:1803.05457v1}.

\bibitem[{Cohan et~al.(2019)Cohan, Ammar, Zuylen, and
  Cady}]{Cohan2019Structural}
Arman Cohan, Waleed Ammar, Madeleine~Van Zuylen, and Field Cady. 2019.
\newblock Structural scaffolds for citation intent classification in scientific
  publications.
\newblock In \emph{NAACL}.

\bibitem[{Conneau et~al.(2017)Conneau, Kiela, Schwenk, Barrault, and
  Bordes}]{conneau-etal-2017-supervised}
Alexis Conneau, Douwe Kiela, Holger Schwenk, Lo{\"\i}c Barrault, and Antoine
  Bordes. 2017.
\newblock \href {https://doi.org/10.18653/v1/D17-1070} {Supervised learning of
  universal sentence representations from natural language inference data}.
\newblock In \emph{Proceedings of EMNLP2017}, pages 670--680, Copenhagen,
  Denmark. Association for Computational Linguistics.

\bibitem[{Conneau et~al.(2018)Conneau, Kruszewski, Lample, Barrault, and
  Baroni}]{conneau-etal-2018-cram}
Alexis Conneau, German Kruszewski, Guillaume Lample, Loic Barrault, and Marco
  Baroni. 2018.
\newblock \href {https://doi.org/10.18653/v1/P18-1198} {What you can cram into
  a single {\$}{\&}!{\#}* vector: Probing sentence embeddings for linguistic
  properties}.
\newblock In \emph{Proceedings of the 56th Annual Meeting of the Association
  for Computational Linguistics (Volume 1: Long Papers)}, pages 2126--2136,
  Melbourne, Australia. Association for Computational Linguistics.

\bibitem[{Coucke et~al.(2018)Coucke, Saade, Ball, Bluche, Caulier, Leroy,
  Doumouro, Gisselbrecht, Caltagirone, Lavril, Primet, and
  Dureau}]{DBLP:journals/corr/abs-1805-10190}
Alice Coucke, Alaa Saade, Adrien Ball, Theodore Bluche, Alexandre Caulier,
  David Leroy, Clement Doumouro, Thibault Gisselbrecht, Francesco Caltagirone,
  Thibaut Lavril, Mael Primet, and Joseph Dureau. 2018.
\newblock \href {http://arxiv.org/abs/1805.10190} {Snips voice platform: an
  embedded spoken language understanding system for private-by-design voice
  interfaces}.
\newblock \emph{CoRR}, abs/1805.10190.

\bibitem[{Cui et~al.(2020)Cui, Wu, Liu, Zhang, and Zhou}]{mutual}
Leyang Cui, Yu~Wu, Shujie Liu, Yue Zhang, and Ming Zhou. 2020.
\newblock Mutual: A dataset for multi-turn dialogue reasoning.
\newblock In \emph{Proceedings of the 58th Conference of the Association for
  Computational Linguistics}. Association for Computational Linguistics.

\bibitem[{Davidson et~al.(2017)Davidson, Warmsley, Macy, and
  Weber}]{hateoffensive}
Thomas Davidson, Dana Warmsley, Michael Macy, and Ingmar Weber. 2017.
\newblock Automated hate speech detection and the problem of offensive
  language.
\newblock In \emph{Proceedings of the 11th International AAAI Conference on Web
  and Social Media}, ICWSM '17, pages 512--515.

\bibitem[{de~Gibert et~al.(2018)de~Gibert, Perez, Garcia-Pablos, and
  Cuadros}]{gibert2018hate}
Ona de~Gibert, Naiara Perez, Aitor Garcia-Pablos, and Montse Cuadros. 2018.
\newblock \href {https://doi.org/10.18653/v1/W18-5102} {{Hate Speech Dataset
  from a White Supremacy Forum}}.
\newblock In \emph{Proceedings of the 2nd Workshop on Abusive Language Online
  ({ALW}2)}, pages 11--20, Brussels, Belgium. Association for Computational
  Linguistics.

\bibitem[{De~Marneffe et~al.(2019)De~Marneffe, Simons, and
  Tonhauser}]{demarneff_simons_tonhauser_2019}
Marie-Catherine De~Marneffe, Mandy Simons, and Judith Tonhauser. 2019.
\newblock The commitmentbank: Investigating projection in naturally occurring
  discourse.
\newblock In \emph{proceedings of Sinn und Bedeutung}, volume~23, pages
  107--124.

\bibitem[{Demszky et~al.(2020)Demszky, Movshovitz-Attias, Ko, Cowen, Nemade,
  and Ravi}]{demszky2020goemotions}
Dorottya Demszky, Dana Movshovitz-Attias, Jeongwoo Ko, Alan Cowen, Gaurav
  Nemade, and Sujith Ravi. 2020.
\newblock {GoEmotions: A Dataset of Fine-Grained Emotions}.
\newblock In \emph{58th Annual Meeting of the Association for Computational
  Linguistics (ACL)}.

\bibitem[{Derczynski et~al.(2017)Derczynski, Nichols, van Erp, and
  Limsopatham}]{derczynski-etal-2017-results}
Leon Derczynski, Eric Nichols, Marieke van Erp, and Nut Limsopatham. 2017.
\newblock \href {https://doi.org/10.18653/v1/W17-4418} {Results of the
  {WNUT}2017 shared task on novel and emerging entity recognition}.
\newblock In \emph{Proceedings of the 3rd Workshop on Noisy User-generated
  Text}, pages 140--147, Copenhagen, Denmark. Association for Computational
  Linguistics.

\bibitem[{Devlin et~al.(2019)Devlin, Chang, Lee, and
  Toutanova}]{devlin-etal-2019-bert}
Jacob Devlin, Ming-Wei Chang, Kenton Lee, and Kristina Toutanova. 2019.
\newblock \href {https://doi.org/10.18653/v1/N19-1423} {{BERT}: Pre-training of
  deep bidirectional transformers for language understanding}.
\newblock In \emph{Proceedings of NAACL2019}, pages 4171--4186, Minneapolis,
  Minnesota. Association for Computational Linguistics.

\bibitem[{Dogan et~al.(2014)Dogan, Leaman, and Lu}]{dougan2014ncbi}
Rezarta~Islamaj Dogan, Robert Leaman, and Zhiyong Lu. 2014.
\newblock Ncbi disease corpus: a resource for disease name recognition and
  concept normalization.
\newblock \emph{Journal of biomedical informatics}, 47:1--10.

\bibitem[{ElSherief et~al.(2021)ElSherief, Ziems, Muchlinski, Anupindi,
  Seybolt, De~Choudhury, and Yang}]{elsherief-etal-2021-latent}
Mai ElSherief, Caleb Ziems, David Muchlinski, Vaishnavi Anupindi, Jordyn
  Seybolt, Munmun De~Choudhury, and Diyi Yang. 2021.
\newblock \href {https://aclanthology.org/2021.emnlp-main.29} {Latent hatred: A
  benchmark for understanding implicit hate speech}.
\newblock In \emph{Proceedings of the 2021 Conference on Empirical Methods in
  Natural Language Processing}, pages 345--363, Online and Punta Cana,
  Dominican Republic. Association for Computational Linguistics.

\bibitem[{Emelin et~al.(2021)Emelin, Bras, Hwang, Forbes, and
  Choi}]{Emelin2021MoralSS}
Denis Emelin, Ronan~Le Bras, Jena~D. Hwang, Maxwell Forbes, and Yejin Choi.
  2021.
\newblock Moral stories: Situated reasoning about norms, intents, actions and
  their consequences.
\newblock \emph{ArXiv}, abs/2012.15738.

\bibitem[{Ethayarajh et~al.(2023)Ethayarajh, Zhang, Wang, and Jurafsky}]{SHP}
Kawin Ethayarajh, Heidi Zhang, Yizhong Wang, and Dan Jurafsky. 2023.
\newblock \href {https://huggingface.co/datasets/stanfordnlp/SHP} {Stanford
  human preferences dataset}.

\bibitem[{Faruqui and Das(2018)}]{faruqui2018identifying}
Manaal Faruqui and Dipanjan Das. 2018.
\newblock \href {http://arxiv.org/abs/1808.09419} {Identifying well-formed
  natural language questions}.

\bibitem[{Feng et~al.(2020)Feng, Liu, Greenspan, and Zhu}]{feng2020exploring}
Ziou~Zheng Feng, Yufei, Quan Liu, Michael Greenspan, and Xiaodan Zhu. 2020.
\newblock Exploring end-to-end differentiable natural logic modeling.
\newblock In \emph{Proceedings of the 28th International Conference on
  Computational Linguistics}, pages 1172--1185.

\bibitem[{Ferreira and Vlachos(2016)}]{Ferreira2016EmergentAN}
William Ferreira and Andreas Vlachos. 2016.
\newblock Emergent: a novel data-set for stance classification.
\newblock In \emph{HLT-NAACL}.

\bibitem[{Fries et~al.(2022)Fries, Weber, Seelam, Altay, Datta, Garda, Kang,
  Su, Kusa, Cahyawijaya, Barth, Ott, Samwald, Bach, Biderman, S{\"a}nger, Wang,
  Callahan, Peri{\~n}{\'a}n, Gigant, Haller, Chim, Posada, Giorgi, Sivaraman,
  P{\`a}mies, Nezhurina, Martin, Cullan, Freidank, Dahlberg, Mishra, Bose,
  Broad, Labrak, Deshmukh, Kiblawi, Singh, Vu, Neeraj, Golde, del Moral, and
  Beilharz}]{fries2022bigbio}
Jason~Alan Fries, Leon Weber, Natasha Seelam, Gabriel Altay, Debajyoti Datta,
  Samuele Garda, Myungsun Kang, Ruisi Su, Wojciech Kusa, Samuel Cahyawijaya,
  Fabio Barth, Simon Ott, Matthias Samwald, Stephen Bach, Stella Biderman,
  Mario S{\"a}nger, Bo~Wang, Alison Callahan, Daniel~Le{\'o}n Peri{\~n}{\'a}n,
  Th{\'e}o Gigant, Patrick Haller, Jenny Chim, Jose~David Posada, John~Michael
  Giorgi, Karthik~Rangasai Sivaraman, Marc P{\`a}mies, Marianna Nezhurina,
  Robert Martin, Michael Cullan, Moritz Freidank, Nathan Dahlberg, Shubhanshu
  Mishra, Shamik Bose, Nicholas~Michio Broad, Yanis Labrak, Shlok~S Deshmukh,
  Sid Kiblawi, Ayush Singh, Minh~Chien Vu, Trishala Neeraj, Jonas Golde,
  Albert~Villanova del Moral, and Benjamin Beilharz. 2022.
\newblock \href {https://openreview.net/forum?id=8lQDn9zTQlW} {Bigbio: A
  framework for data-centric biomedical natural language processing}.
\newblock In \emph{Thirty-sixth Conference on Neural Information Processing
  Systems Datasets and Benchmarks Track}.

\bibitem[{George and Mamidi(2020)}]{george2020conversational}
Elizabeth~Jasmi George and Radhika Mamidi. 2020.
\newblock Conversational implicatures in english dialogue: Annotated dataset.
\newblock \emph{Procedia Computer Science}, 171:2316--2323.

\bibitem[{Ghosal et~al.(2022)Ghosal, Shen, Majumder, Mihalcea, and
  Poria}]{ghosal2022cicero}
Deepanway Ghosal, Siqi Shen, Navonil Majumder, Rada Mihalcea, and Soujanya
  Poria. 2022.
\newblock Cicero: A dataset for contextualized commonsense inference in
  dialogues.
\newblock In \emph{Proceedings of the 60th Annual Meeting of the Association
  for Computational Linguistics (Volume 1: Long Papers)}, pages 5010--5028.

\bibitem[{Godfrey et~al.(1992)Godfrey, Holliman, and
  McDaniel}]{Godfrey:1992:STS:1895550.1895693}
John~J. Godfrey, Edward~C. Holliman, and Jane McDaniel. 1992.
\newblock \href {http://dl.acm.org/citation.cfm?id=1895550.1895693}
  {Switchboard: Telephone speech corpus for research and development}.
\newblock In \emph{Proceedings of the 1992 IEEE International Conference on
  Acoustics, Speech and Signal Processing - Volume 1}, ICASSP'92, pages
  517--520, Washington, DC, USA. IEEE Computer Society.

\bibitem[{Gorrell et~al.(2019)Gorrell, Kochkina, Liakata, Aker, Zubiaga,
  Bontcheva, and Derczynski}]{gorrell-etal-2019-semeval}
Genevieve Gorrell, Elena Kochkina, Maria Liakata, Ahmet Aker, Arkaitz Zubiaga,
  Kalina Bontcheva, and Leon Derczynski. 2019.
\newblock \href {https://doi.org/10.18653/v1/S19-2147} {{S}em{E}val-2019 task
  7: {R}umour{E}val, determining rumour veracity and support for rumours}.
\newblock In \emph{Proceedings of the 13th International Workshop on Semantic
  Evaluation}, pages 845--854, Minneapolis, Minnesota, USA. Association for
  Computational Linguistics.

\bibitem[{Gurulingappa et~al.(2012)Gurulingappa, Rajput, Roberts, Fluck,
  Hofmann-Apitius, and Toldo}]{GURULINGAPPA2012885}
Harsha Gurulingappa, Abdul~Mateen Rajput, Angus Roberts, Juliane Fluck, Martin
  Hofmann-Apitius, and Luca Toldo. 2012.
\newblock \href {https://doi.org/https://doi.org/10.1016/j.jbi.2012.04.008}
  {Development of a benchmark corpus to support the automatic extraction of
  drug-related adverse effects from medical case reports}.
\newblock \emph{Journal of Biomedical Informatics}, 45(5):885 -- 892.
\newblock Text Mining and Natural Language Processing in Pharmacogenomics.

\bibitem[{Gusev and Tikhonov(2021)}]{gusev2021headlinecause}
Ilya Gusev and Alexey Tikhonov. 2021.
\newblock \href {http://arxiv.org/abs/2108.12626} {Headlinecause: A dataset of
  news headlines for detecting casualties}.

\bibitem[{Habernal et~al.(2018)Habernal, Wachsmuth, Gurevych, and
  Stein}]{Habernal.et.al.2018.NAACL.ARCT}
Ivan Habernal, Henning Wachsmuth, Iryna Gurevych, and Benno Stein. 2018.
\newblock \href {http://aclweb.org/anthology/N18-1175} {The argument reasoning
  comprehension task: Identification and reconstruction of implicit warrants}.
\newblock In \emph{Proceedings of the 2018 Conference of the North American
  Chapter of the Association for Computational Linguistics: Human Language
  Technologies, Volume 1 (Long Papers)}, pages 1930--1940, New Orleans,
  Louisiana. Association for Computational Linguistics.

\bibitem[{Han et~al.(2022)Han, Schoelkopf, Zhao, Qi, Riddell, Benson, Sun,
  Zubova, Qiao, Burtell, Peng, Fan, Liu, Wong, Sailor, Ni, Nan, Kasai, Yu,
  Zhang, Joty, Fabbri, Kryscinski, Lin, Xiong, and Radev}]{han2022folio}
Simeng Han, Hailey Schoelkopf, Yilun Zhao, Zhenting Qi, Martin Riddell, Luke
  Benson, Lucy Sun, Ekaterina Zubova, Yujie Qiao, Matthew Burtell, David Peng,
  Jonathan Fan, Yixin Liu, Brian Wong, Malcolm Sailor, Ansong Ni, Linyong Nan,
  Jungo Kasai, Tao Yu, Rui Zhang, Shafiq Joty, Alexander~R. Fabbri, Wojciech
  Kryscinski, Xi~Victoria Lin, Caiming Xiong, and Dragomir Radev. 2022.
\newblock \href {https://arxiv.org/abs/2209.00840} {Folio: Natural language
  reasoning with first-order logic}.
\newblock \emph{arXiv preprint arXiv:2209.00840}.

\bibitem[{He et~al.(2021)He, Gao, and Chen}]{he2021debertav3}
Pengcheng He, Jianfeng Gao, and Weizhu Chen. 2021.
\newblock Debertav3: Improving deberta using electra-style pre-training with
  gradient-disentangled embedding sharing.
\newblock \emph{arXiv preprint arXiv:2111.09543}.

\bibitem[{Hendrickx et~al.(2010)Hendrickx, Kim, Kozareva, Nakov, O~Seaghdha,
  Pado, Pennacchiotti, Romano, and Szpakowicz}]{hendrickx-etal-2010-semeval}
Iris Hendrickx, Su~Nam Kim, Zornitsa Kozareva, Preslav Nakov, Diarmuid
  O~Seaghdha, Sebastian Pado, Marco Pennacchiotti, Lorenza Romano, and Stan
  Szpakowicz. 2010.
\newblock \href {https://www.aclweb.org/anthology/S10-1006} {{S}em{E}val-2010
  task 8: Multi-way classification of semantic relations between pairs of
  nominals}.
\newblock In \emph{Proceedings of the 5th International Workshop on Semantic
  Evaluation}, pages 33--38, Uppsala, Sweden. Association for Computational
  Linguistics.

\bibitem[{Hendrycks et~al.(2020)Hendrycks, Burns, Basart, Critch, Li, Song, and
  Steinhardt}]{hendrycks2020ethics}
Dan Hendrycks, Collin Burns, Steven Basart, Andrew Critch, Jerry Li, Dawn Song,
  and Jacob Steinhardt. 2020.
\newblock Aligning ai with shared human values.
\newblock \emph{arXiv preprint arXiv:2008.02275}.

\bibitem[{Hendrycks et~al.(2021)Hendrycks, Burns, Basart, Zou, Mazeika, Song,
  and Steinhardt}]{hendryckstest2021}
Dan Hendrycks, Collin Burns, Steven Basart, Andy Zou, Mantas Mazeika, Dawn
  Song, and Jacob Steinhardt. 2021.
\newblock Measuring massive multitask language understanding.
\newblock \emph{Proceedings of the International Conference on Learning
  Representations (ICLR)}.

\bibitem[{Hossain et~al.(2019)Hossain, Krumm, and Gamon}]{hossain2019president}
Nabil Hossain, John Krumm, and Michael Gamon. 2019.
\newblock " president vows to cut< taxes> hair": Dataset and analysis of
  creative text editing for humorous headlines.
\newblock \emph{arXiv preprint arXiv:1906.00274}.

\bibitem[{Huang et~al.(2019)Huang, Le~Bras, Bhagavatula, and
  Choi}]{huang-etal-2019-cosmos}
Lifu Huang, Ronan Le~Bras, Chandra Bhagavatula, and Yejin Choi. 2019.
\newblock \href {https://doi.org/10.18653/v1/D19-1243} {Cosmos {QA}: Machine
  reading comprehension with contextual commonsense reasoning}.
\newblock In \emph{Proceedings of the 2019 Conference on Empirical Methods in
  Natural Language Processing and the 9th International Joint Conference on
  Natural Language Processing (EMNLP-IJCNLP)}, pages 2391--2401, Hong Kong,
  China. Association for Computational Linguistics.

\bibitem[{Huang et~al.(2020)Huang, Liu, and
  Bowman}]{huang2020cnligeneralization}
William Huang, Haokun Liu, and Samuel~R. Bowman. 2020.
\newblock Counterfactually-augmented {SNLI} training data does not yield better
  generalization than unaugmented data.
\newblock In \emph{Proceedings of the 2020 EMNLP Workshop on Insights from
  Negative Results in NLP}. The Association for Computational Linguistics.

\bibitem[{huggingface(2020)}]{huggingface:dataset}
Inc. huggingface. 2020.
\newblock A great new dataset.

\bibitem[{Jereti~c et~al.(2020)Jereti~c, Warstadt, Bhooshan, and
  Williams}]{jeretic-etal-2020-natural}
Paloma Jereti~c, Alex Warstadt, Suvrat Bhooshan, and Adina Williams. 2020.
\newblock \href {https://doi.org/10.18653/v1/2020.acl-main.768} {Are natural
  language inference models {IMPPRESsive}? {L}earning {IMPlicature} and
  {PRESupposition}}.
\newblock In \emph{Proceedings of the 58th Annual Meeting of the Association
  for Computational Linguistics}, pages 8690--8705, Online. Association for
  Computational Linguistics.

\bibitem[{Johannes~Welbl(2017)}]{SciQ}
Matt~Gardner Johannes~Welbl, Nelson F.~Liu. 2017.
\newblock Crowdsourcing multiple choice science questions.

\bibitem[{Joulin et~al.(2016)Joulin, Grave, Bojanowski, Douze, J{\'e}gou, and
  Mikolov}]{joulin2016fasttext}
Armand Joulin, Edouard Grave, Piotr Bojanowski, Matthijs Douze, H{\'e}rve
  J{\'e}gou, and Tomas Mikolov. 2016.
\newblock Fasttext.zip: Compressing text classification models.
\newblock \emph{arXiv preprint arXiv:1612.03651}.

\bibitem[{Kaushik et~al.(2020)Kaushik, Hovy, and Lipton}]{kaushik2020learning}
Divyansh Kaushik, Eduard Hovy, and Zachary~C Lipton. 2020.
\newblock Learning the difference that makes a difference with counterfactually
  augmented data.
\newblock \emph{International Conference on Learning Representations (ICLR)}.

\bibitem[{Kavumba et~al.(2019)Kavumba, Inoue, Heinzerling, Singh, Reisert, and
  Inui}]{kavumba-etal-2019-choosing}
Pride Kavumba, Naoya Inoue, Benjamin Heinzerling, Keshav Singh, Paul Reisert,
  and Kentaro Inui. 2019.
\newblock \href {https://doi.org/10.18653/v1/D19-6004} {When choosing plausible
  alternatives, clever hans can be clever}.
\newblock In \emph{Proceedings of the First Workshop on Commonsense Inference
  in Natural Language Processing}, pages 33--42, Hong Kong, China. Association
  for Computational Linguistics.

\bibitem[{Kejriwal and Shen(2020)}]{Kejriwal2020DoFC}
Mayank Kejriwal and Ke~Shen. 2020.
\newblock Do fine-tuned commonsense language models really generalize?
\newblock \emph{ArXiv}, abs/2011.09159.

\bibitem[{Khashabi et~al.(2020{\natexlab{a}})Khashabi, Khot, and
  Sabhwaral}]{khashabi2020naturalperturbations}
D.~Khashabi, T.~Khot, and A.~Sabhwaral. 2020{\natexlab{a}}.
\newblock Natural perturbation for robust question answering.
\newblock \emph{arXiv preprint}.

\bibitem[{Khashabi et~al.(2020{\natexlab{b}})Khashabi, Min, Khot, Sabhwaral,
  Tafjord, Clark, and Hajishirzi}]{2020unifiedqa}
D.~Khashabi, S.~Min, T.~Khot, A.~Sabhwaral, O.~Tafjord, P.~Clark, and
  H.~Hajishirzi. 2020{\natexlab{b}}.
\newblock Unifiedqa: Crossing format boundaries with a single qa system.

\bibitem[{Khashabi et~al.(2018)Khashabi, Chaturvedi, Roth, Upadhyay, and
  Roth}]{MultiRC2018}
Daniel Khashabi, Snigdha Chaturvedi, Michael Roth, Shyam Upadhyay, and Dan
  Roth. 2018.
\newblock Looking beyond the surface:a challenge set for reading comprehension
  over multiple sentences.
\newblock In \emph{Proceedings of North American Chapter of the Association for
  Computational Linguistics (NAACL)}.

\bibitem[{Khot et~al.(2020)Khot, Clark, Guerquin, Jansen, and
  Sabharwal}]{allenai:qasc}
Tushar Khot, Peter Clark, Michal Guerquin, Peter Jansen, and Ashish Sabharwal.
  2020.
\newblock Qasc: A dataset for question answering via sentence composition.
\newblock \emph{arXiv:1910.11473v2}.

\bibitem[{Khot et~al.(2018)Khot, Sabharwal, and Clark}]{scitail}
Tushar Khot, Ashish Sabharwal, and Peter Clark. 2018.
\newblock {SciTail}: A textual entailment dataset from science question
  answering.
\newblock In \emph{AAAI}.

\bibitem[{Kim et~al.(2004)Kim, Ohta, Tsuruoka, Tateisi, and
  Collier}]{kim2004introduction}
Jin-Dong Kim, Tomoko Ohta, Yoshimasa Tsuruoka, Yuka Tateisi, and Nigel Collier.
  2004.
\newblock Introduction to the bio-entity recognition task at jnlpba.
\newblock In \emph{Proceedings of the international joint workshop on natural
  language processing in biomedicine and its applications}, pages 70--75.
  Citeseer.

\bibitem[{Kotonya and Toni(2020)}]{kotonya-toni-2020-explainable}
Neema Kotonya and Francesca Toni. 2020.
\newblock \href {https://www.aclweb.org/anthology/2020.emnlp-main.623}
  {Explainable automated fact-checking for public health claims}.
\newblock In \emph{Proceedings of the 2020 Conference on Empirical Methods in
  Natural Language Processing (EMNLP)}, pages 7740--7754, Online. Association
  for Computational Linguistics.

\bibitem[{Laban and Bandarkar(2021)}]{Laban2021NewsHG}
Philippe Laban and Lucas Bandarkar. 2021.
\newblock News headline grouping as a challenging nlu task.
\newblock In \emph{NAACL 2021}. Association for Computational Linguistics.

\bibitem[{Lahiri(2015)}]{DBLP:journals/corr/Lahiri15}
Shibamouli Lahiri. 2015.
\newblock \href {http://arxiv.org/abs/1506.02306} {{SQUINKY! A Corpus of
  Sentence-level Formality, Informativeness, and Implicature}}.
\newblock \emph{CoRR}, abs/1506.02306.

\bibitem[{Lai et~al.(2017)Lai, Xie, Liu, Yang, and Hovy}]{lai2017large}
Guokun Lai, Qizhe Xie, Hanxiao Liu, Yiming Yang, and Eduard Hovy. 2017.
\newblock Race: Large-scale reading comprehension dataset from examinations.
\newblock \emph{arXiv preprint arXiv:1704.04683}.

\bibitem[{Lau et~al.(2015)Lau, Clark, and Lappin}]{lau-etal-2015-unsupervised}
Jey~Han Lau, Alexander Clark, and Shalom Lappin. 2015.
\newblock \href {https://doi.org/10.3115/v1/P15-1156} {Unsupervised prediction
  of acceptability judgements}.
\newblock In \emph{Proceedings of the 53rd Annual Meeting of the Association
  for Computational Linguistics and the 7th International Joint Conference on
  Natural Language Processing (Volume 1: Long Papers)}, pages 1618--1628,
  Beijing, China. Association for Computational Linguistics.

\bibitem[{Laurer et~al.(2022)Laurer, Atteveldt, Casas, and
  Welbers}]{laurer2022less}
Moritz Laurer, W~v Atteveldt, Andreu Casas, and Kasper Welbers. 2022.
\newblock Less annotating, more classifying--addressing the data scarcity issue
  of supervised machine learning with deep transfer learning and bert-nli.

\bibitem[{Leech and Weisser(2003)}]{leech2003generic}
Geoffrey Leech and Martin Weisser. 2003.
\newblock Generic speech act annotation for task-oriented dialogues.
\newblock In \emph{Proceedings of the corpus linguistics 2003 conference},
  volume~16, pages 441--446. Lancaster: Lancaster University.

\bibitem[{Lehmann et~al.(2015)Lehmann, Isele, Jakob, Jentzsch, Kontokostas,
  Mendes, Hellmann, Morsey, Van~Kleef, Auer et~al.}]{lehmann2015dbpedia}
Jens Lehmann, Robert Isele, Max Jakob, Anja Jentzsch, Dimitris Kontokostas,
  Pablo~N Mendes, Sebastian Hellmann, Mohamed Morsey, Patrick Van~Kleef, Soren
  Auer, et~al. 2015.
\newblock Dbpedia--a large-scale, multilingual knowledge base extracted from
  wikipedia.
\newblock \emph{Semantic web}, 6(2):167--195.

\bibitem[{Li and Roth(2002)}]{li-roth-2002-learning}
Xin Li and Dan Roth. 2002.
\newblock \href {https://www.aclweb.org/anthology/C02-1150} {Learning question
  classifiers}.
\newblock In \emph{{COLING} 2002: The 19th International Conference on
  Computational Linguistics}.

\bibitem[{Li et~al.(2017)Li, Su, Shen, Li, Cao, and Niu}]{li2017dailydialog}
Yanran Li, Hui Su, Xiaoyu Shen, Wenjie Li, Ziqiang Cao, and Shuzi Niu. 2017.
\newblock Dailydialog: A manually labelled multi-turn dialogue dataset.
\newblock In \emph{Proceedings of The 8th International Joint Conference on
  Natural Language Processing (IJCNLP 2017)}.

\bibitem[{Liang et~al.(2019)Liang, Li, and Yin}]{pmlr-v101-liang19a}
Yichan Liang, Jianheng Li, and Jian Yin. 2019.
\newblock A new multi-choice reading comprehension dataset for curriculum
  learning.
\newblock In \emph{Proceedings of The Eleventh Asian Conference on Machine
  Learning}, pages 742--757.

\bibitem[{Lin et~al.(2020)Lin, Lee, Khanna, and Ren}]{lin2020numersense}
Bill~Yuchen Lin, Seyeon Lee, Rahul Khanna, and Xiang Ren. 2020.
\newblock Birds have four legs?! numersense: Probing numerical commonsense
  knowledge of pre-trained language models.
\newblock In \emph{Proceedings of EMNLP}.
\newblock To appear.

\bibitem[{Lin et~al.(2021{\natexlab{a}})Lin, Wu, Yang, Lee, and
  Ren}]{lin-etal-2021-riddlesense}
Bill~Yuchen Lin, Ziyi Wu, Yichi Yang, Dong-Ho Lee, and Xiang Ren.
  2021{\natexlab{a}}.
\newblock Riddlesense: Reasoning about riddle questions featuring linguistic
  creativity and commonsense knowledge.

\bibitem[{Lin et~al.(2021{\natexlab{b}})Lin, Hilton, and
  Evans}]{lin2021truthfulqa}
Stephanie Lin, Jacob Hilton, and Owain Evans. 2021{\natexlab{b}}.
\newblock \href {http://arxiv.org/abs/2109.07958} {Truthfulqa: Measuring how
  models mimic human falsehoods}.

\bibitem[{Lippi et~al.(2019)Lippi, Paka, Contissa, Lagioia, Micklitz, Sartor,
  and Torroni}]{lippi-etal-2019-claudette}
Marco Lippi, Przemysaw Paka, Giuseppe Contissa, Francesca Lagioia,
  Hans-Wolfgang Micklitz, Giovanni Sartor, and Paolo Torroni. 2019.
\newblock \href {https://doi.org/10.1007/s10506-019-09243-2} {{CLAUDETTE}: an
  automated detector of potentially unfair clauses in online terms of service}.
\newblock \emph{Artificial Intelligence and Law}, pages 117--139.

\bibitem[{Liu et~al.(2022)Liu, Swayamdipta, Smith, and
  Choi}]{liu-etal-2022-wanli}
Alisa Liu, Swabha Swayamdipta, Noah~A. Smith, and Yejin Choi. 2022.
\newblock \href {https://arxiv.org/pdf/2201.05955} {Wanli: Worker and ai
  collaboration for natural language inference dataset creation}.

\bibitem[{Liu et~al.(2023)Liu, Wu, Michael, Suhr, West, Koller, Swayamdipta,
  Smith, and Choi}]{liu-etal-2023-afraid}
Alisa Liu, Zhaofeng Wu, Julian Michael, Alane Suhr, Peter West, Alexander
  Koller, Swabha Swayamdipta, Noah~A. Smith, and Yejin Choi. 2023.
\newblock \href {https://arxiv.org/abs/2304.14399} {We're afraid language
  models aren't modeling ambiguity}.

\bibitem[{Liu et~al.(2020)Liu, Cui, Liu, Huang, Wang, and
  Zhang}]{liu2020logiqa}
Jian Liu, Leyang Cui, Hanmeng Liu, Dandan Huang, Yile Wang, and Yue Zhang.
  2020.
\newblock Logiqa: A challenge dataset for machine reading comprehension with
  logical reasoning.
\newblock \emph{arXiv preprint arXiv:2007.08124}.

\bibitem[{Louis et~al.(2020)Louis, Roth, and Radlinski}]{louis_emnlp2020}
Annie Louis, Dan Roth, and Filip Radlinski. 2020.
\newblock I'd rather just go to bed: Understanding indirect answers.
\newblock In \emph{Proceedings of the 2020 Conference on Empirical Methods in
  Natural Language Processing}.

\bibitem[{Maas et~al.(2011)Maas, Daly, Pham, Huang, Ng, and
  Potts}]{maas-EtAl:2011:ACL-HLT2011}
Andrew~L. Maas, Raymond~E. Daly, Peter~T. Pham, Dan Huang, Andrew~Y. Ng, and
  Christopher Potts. 2011.
\newblock \href {http://www.aclweb.org/anthology/P11-1015} {Learning word
  vectors for sentiment analysis}.
\newblock In \emph{Proceedings of the 49th Annual Meeting of the Association
  for Computational Linguistics: Human Language Technologies}, pages 142--150,
  Portland, Oregon, USA. Association for Computational Linguistics.

\bibitem[{Malo et~al.(2014)Malo, Sinha, Korhonen, Wallenius, and
  Takala}]{Malo2014GoodDO}
P.~Malo, A.~Sinha, P.~Korhonen, J.~Wallenius, and P.~Takala. 2014.
\newblock Good debt or bad debt: Detecting semantic orientations in economic
  texts.
\newblock \emph{Journal of the Association for Information Science and
  Technology}, 65.

\bibitem[{Marelli et~al.(2014)Marelli, Menini, Baroni, Bentivogli, Bernardi,
  and Zamparelli}]{marelli-etal-2014-sick}
Marco Marelli, Stefano Menini, Marco Baroni, Luisa Bentivogli, Raffaella
  Bernardi, and Roberto Zamparelli. 2014.
\newblock \href
  {http://www.lrec-conf.org/proceedings/lrec2014/pdf/363_Paper.pdf} {A {SICK}
  cure for the evaluation of compositional distributional semantic models}.
\newblock In \emph{Proceedings of the Ninth International Conference on
  Language Resources and Evaluation ({LREC}'14)}, pages 216--223, Reykjavik,
  Iceland. European Language Resources Association (ELRA).

\bibitem[{McAuley and Leskovec(2013)}]{mcauley2013hidden}
Julian McAuley and Jure Leskovec. 2013.
\newblock Hidden factors and hidden topics: understanding rating dimensions
  with review text.
\newblock In \emph{Proceedings of the 7th ACM conference on Recommender
  systems}, pages 165--172.

\bibitem[{McCoy et~al.(2019)McCoy, Pavlick, and
  Linzen}]{DBLP:journals/corr/abs-1902-01007}
R.~Thomas McCoy, Ellie Pavlick, and Tal Linzen. 2019.
\newblock \href {http://arxiv.org/abs/1902.01007} {Right for the wrong reasons:
  Diagnosing syntactic heuristics in natural language inference}.
\newblock \emph{CoRR}, abs/1902.01007.

\bibitem[{McCreery et~al.(2020)McCreery, Katariya, Kannan, Chablani, and
  Amatriain}]{mccreery2020effective}
Clara~H. McCreery, Namit Katariya, Anitha Kannan, Manish Chablani, and Xavier
  Amatriain. 2020.
\newblock \href {http://arxiv.org/abs/2008.13546} {Effective transfer learning
  for identifying similar questions: Matching user questions to covid-19 faqs}.

\bibitem[{McKeown et~al.(2011)McKeown, Valstar, Cowie, Pantic, and
  Schroder}]{mckeown2011semaine}
Gary McKeown, Michel Valstar, Roddy Cowie, Maja Pantic, and Marc Schroder.
  2011.
\newblock The semaine database: Annotated multimodal records of emotionally
  colored conversations between a person and a limited agent.
\newblock \emph{IEEE transactions on affective computing}, 3(1):5--17.

\bibitem[{Mihaylov et~al.(2018)Mihaylov, Clark, Khot, and
  Sabharwal}]{OpenBookQA2018}
Todor Mihaylov, Peter Clark, Tushar Khot, and Ashish Sabharwal. 2018.
\newblock Can a suit of armor conduct electricity? a new dataset for open book
  question answering.
\newblock In \emph{EMNLP}.

\bibitem[{Min et~al.(2020)Min, McCoy, Das, Pitler, and
  Linzen}]{min-etal-2020-syntactic}
Junghyun Min, R.~Thomas McCoy, Dipanjan Das, Emily Pitler, and Tal Linzen.
  2020.
\newblock \href {https://doi.org/10.18653/v1/2020.acl-main.212} {Syntactic data
  augmentation increases robustness to inference heuristics}.
\newblock In \emph{Proceedings of the 58th Annual Meeting of the Association
  for Computational Linguistics}, pages 2339--2352, Online. Association for
  Computational Linguistics.

\bibitem[{Mirzaee et~al.(2021)Mirzaee, Rajaby~Faghihi, Ning, and
  Kordjamshidi}]{mirzaee-etal-2021-spartqa}
Roshanak Mirzaee, Hossein Rajaby~Faghihi, Qiang Ning, and Parisa Kordjamshidi.
  2021.
\newblock \href {https://doi.org/10.18653/v1/2021.naacl-main.364} {{SPARTQA}: A
  textual question answering benchmark for spatial reasoning}.
\newblock In \emph{Proceedings of the 2021 Conference of the North American
  Chapter of the Association for Computational Linguistics: Human Language
  Technologies}, pages 4582--4598, Online. Association for Computational
  Linguistics.

\bibitem[{Mishra et~al.(2022)Mishra, Khashabi, Baral, and
  Hajishirzi}]{naturalinstructions}
Swaroop Mishra, Daniel Khashabi, Chitta Baral, and Hannaneh Hajishirzi. 2022.
\newblock Cross-task generalization via natural language crowdsourcing
  instructions.
\newblock In \emph{ACL}.

\bibitem[{Mollas et~al.(2020)Mollas, Chrysopoulou, Karlos, and
  Tsoumakas}]{mollas2020ethos}
Ioannis Mollas, Zoe Chrysopoulou, Stamatis Karlos, and Grigorios Tsoumakas.
  2020.
\newblock \href {http://arxiv.org/abs/2006.08328} {Ethos: an online hate speech
  detection dataset}.

\bibitem[{Nakano et~al.(2021)Nakano, Hilton, Balaji, Wu, Ouyang, Kim, Hesse,
  Jain, Kosaraju, Saunders, Jiang, Cobbe, Eloundou, Krueger, Button, Knight,
  Chess, and Schulman}]{nakano2021webgpt}
Reiichiro Nakano, Jacob Hilton, Suchir Balaji, Jeff Wu, Long Ouyang, Christina
  Kim, Christopher Hesse, Shantanu Jain, Vineet Kosaraju, William Saunders,
  Xu~Jiang, Karl Cobbe, Tyna Eloundou, Gretchen Krueger, Kevin Button, Matthew
  Knight, Benjamin Chess, and John Schulman. 2021.
\newblock Webgpt: Browser-assisted question-answering with human feedback.
\newblock In \emph{arXiv}.

\bibitem[{Nangia et~al.(2020)Nangia, Vania, Bhalerao, and
  Bowman}]{nangia-etal-2020-crows}
Nikita Nangia, Clara Vania, Rasika Bhalerao, and Samuel~R. Bowman. 2020.
\newblock \href {https://doi.org/10.18653/v1/2020.emnlp-main.154}
  {{C}row{S}-pairs: A challenge dataset for measuring social biases in masked
  language models}.
\newblock In \emph{Proceedings of the 2020 Conference on Empirical Methods in
  Natural Language Processing (EMNLP)}, pages 1953--1967, Online. Association
  for Computational Linguistics.

\bibitem[{Nie et~al.(2020)Nie, Williams, Dinan, Bansal, Weston, and
  Kiela}]{nie2019adversarial}
Yixin Nie, Adina Williams, Emily Dinan, Mohit Bansal, Jason Weston, and Douwe
  Kiela. 2020.
\newblock Adversarial nli: A new benchmark for natural language understanding.
\newblock In \emph{Proceedings of the 58th Annual Meeting of the Association
  for Computational Linguistics}. Association for Computational Linguistics.

\bibitem[{ONeill et~al.(2021)ONeill, Rozenshtein, Kiryo, Kubota, and
  Bollegala}]{oneill2021i}
James ONeill, Polina Rozenshtein, Ryuichi Kiryo, Motoko Kubota, and Danushka
  Bollegala. 2021.
\newblock \href {http://arxiv.org/abs/2104.06893} {I wish i would have loved
  this one, but i didn't -- a multilingual dataset for counterfactual detection
  in product reviews}.

\bibitem[{Oraby et~al.(2016)Oraby, Harrison, Reed, Hernandez, Riloff, and
  Walker}]{OrabySarc}
Shereen Oraby, Vrindavan Harrison, Lena Reed, Ernesto Hernandez, Ellen Riloff,
  and Marilyn Walker. 2016.
\newblock \href {https://doi.org/10.18653/v1/W16-3604} {Creating and
  characterizing a diverse corpus of sarcasm in dialogue}.
\newblock In \emph{Proceedings of the 17th Annual Meeting of the Special
  Interest Group on Discourse and Dialogue}, pages 31--41. Association for
  Computational Linguistics.

\bibitem[{Pafilis et~al.(2013)Pafilis, Frankild, Fanini, Faulwetter, Pavloudi,
  Vasileiadou, Arvanitidis, and Jensen}]{pafilis2013species}
Evangelos Pafilis, Sune~P Frankild, Lucia Fanini, Sarah Faulwetter, Christina
  Pavloudi, Aikaterini Vasileiadou, Christos Arvanitidis, and Lars~Juhl Jensen.
  2013.
\newblock The species and organisms resources for fast and accurate
  identification of taxonomic names in text.
\newblock \emph{PloS one}, 8(6):e65390.

\bibitem[{Pal et~al.(2022)Pal, Umapathi, and Sankarasubbu}]{pmlr-v174-pal22a}
Ankit Pal, Logesh~Kumar Umapathi, and Malaikannan Sankarasubbu. 2022.
\newblock \href {https://proceedings.mlr.press/v174/pal22a.html} {Medmcqa: A
  large-scale multi-subject multi-choice dataset for medical domain question
  answering}.
\newblock In \emph{Proceedings of the Conference on Health, Inference and
  Learning}, volume 174 of \emph{Proceedings of Machine Learning Research},
  pages 248--260. PMLR.

\bibitem[{Pang and Lee(2005)}]{Pang+Lee:05a}
Bo~Pang and Lillian Lee. 2005.
\newblock Seeing stars: Exploiting class relationships for sentiment
  categorization with respect to rating scales.
\newblock In \emph{Proceedings of the ACL}.

\bibitem[{Park and Cardie(2014)}]{park2014identifying}
Joonsuk Park and Claire Cardie. 2014.
\newblock Identifying appropriate support for propositions in online user
  comments.
\newblock In \emph{Proceedings of the first workshop on argumentation mining},
  pages 29--38.

\bibitem[{Parrish et~al.(2021)Parrish, Huang, Agha, Lee, Nangia, Warstadt,
  Aggarwal, Allaway, Linzen, and Bowman}]{parrish-etal-2021-putting-linguist}
Alicia Parrish, William Huang, Omar Agha, Soo-Hwan Lee, Nikita Nangia, Alexia
  Warstadt, Karmanya Aggarwal, Emily Allaway, Tal Linzen, and Samuel~R. Bowman.
  2021.
\newblock \href {https://doi.org/10.18653/v1/2021.findings-emnlp.421} {Does
  putting a linguist in the loop improve {NLU} data collection?}
\newblock In \emph{Findings of the Association for Computational Linguistics:
  EMNLP 2021}, pages 4886--4901, Punta Cana, Dominican Republic. Association
  for Computational Linguistics.

\bibitem[{Peng et~al.(2019)Peng, Yan, and Lu}]{peng2019transfer}
Yifan Peng, Shankai Yan, and Zhiyong Lu. 2019.
\newblock Transfer learning in biomedical natural language processing: An
  evaluation of bert and elmo on ten benchmarking datasets.
\newblock In \emph{Proceedings of the 2019 Workshop on Biomedical Natural
  Language Processing (BioNLP 2019)}.

\bibitem[{Perez et~al.(2022)Perez, Ringer, Lukoit, Nguyen, Chen, Heiner,
  Pettit, Olsson, Kundu, Kadavath, Jones, Chen, Mann, Israel, Seethor,
  McKinnon, Olah, Yan, Amodei, Amodei, Drain, Li, Tran-Johnson, Khundadze,
  Kernion, Landis, Kerr, Mueller, Hyun, Landau, Ndousse, Goldberg, Lovitt,
  Lucas, Sellitto, Zhang, Kingsland, Elhage, Joseph, Mercado, DasSarma, Rausch,
  Larson, McCandlish, Johnston, Kravec, E, Lanham, Telleen-Lawton, Brown,
  Henighan, Hume, Bai, Hatfield-Dodds, Clark, Bowman, Askell, Grosse,
  Hernandez, Ganguli, Hubinger, Schiefer, and Kaplan}]{perez2022discovering}
Ethan Perez, Sam Ringer, Kamil Lukoit, Karina Nguyen, Edwin Chen, Scott Heiner,
  Craig Pettit, Catherine Olsson, Sandipan Kundu, Saurav Kadavath, Andy Jones,
  Anna Chen, Ben Mann, Brian Israel, Bryan Seethor, Cameron McKinnon,
  Christopher Olah, Da~Yan, Daniela Amodei, Dario Amodei, Dawn Drain, Dustin
  Li, Eli Tran-Johnson, Guro Khundadze, Jackson Kernion, James Landis, Jamie
  Kerr, Jared Mueller, Jeeyoon Hyun, Joshua Landau, Kamal Ndousse, Landon
  Goldberg, Liane Lovitt, Martin Lucas, Michael Sellitto, Miranda Zhang, Neerav
  Kingsland, Nelson Elhage, Nicholas Joseph, Noem Mercado, Nova DasSarma,
  Oliver Rausch, Robin Larson, Sam McCandlish, Scott Johnston, Shauna Kravec,
  Sheer E, Tamera Lanham, Timothy Telleen-Lawton, Tom Brown, Tom Henighan,
  Tristan Hume, Yuntao Bai, Zac Hatfield-Dodds, Jack Clark, Samuel~R. Bowman,
  Amanda Askell, Roger Grosse, Danny Hernandez, Deep Ganguli, Evan Hubinger,
  Nicholas Schiefer, and Jared Kaplan. 2022.
\newblock \href {https://doi.org/10.48550/ARXIV.2212.09251} {Discovering
  language model behaviors with model-written evaluations}.

\bibitem[{Pham et~al.(2022)Pham, Yoon, Bui, and Nguyen}]{pham2022PiC}
Thang~M Pham, Seunghyun Yoon, Trung Bui, and Anh Nguyen. 2022.
\newblock Pic: A phrase-in-context dataset for phrase understanding and
  semantic search.
\newblock \emph{arXiv preprint arXiv:2207.09068}.

\bibitem[{Pilehvar and ose
  Camacho-Collados(2018)}]{DBLP:journals/corr/abs-1808-09121}
Mohammad~Taher Pilehvar and ose Camacho-Collados. 2018.
\newblock \href {http://arxiv.org/abs/1808.09121} {Wic: 10, 000 example pairs
  for evaluating context-sensitive representations}.
\newblock \emph{CoRR}, abs/1808.09121.

\bibitem[{Poliak et~al.(2018)Poliak, Haldar, Rudinger, Hu, Pavlick, White, and
  Van~Durme}]{poliak-etal-2018-collecting}
Adam Poliak, Aparajita Haldar, Rachel Rudinger, J.~Edward Hu, Ellie Pavlick,
  Aaron~Steven White, and Benjamin Van~Durme. 2018.
\newblock \href {https://doi.org/10.18653/v1/D18-1007} {Collecting diverse
  natural language inference problems for sentence representation evaluation}.
\newblock In \emph{Proceedings of the 2018 Conference on Empirical Methods in
  Natural Language Processing}, pages 67--81, Brussels, Belgium. Association
  for Computational Linguistics.

\bibitem[{Potts et~al.(2020)Potts, Wu, Geiger, and
  Kiela}]{potts-etal-2020-dynasent}
Christopher Potts, Zhengxuan Wu, Atticus Geiger, and Douwe Kiela. 2020.
\newblock \href {https://arxiv.org/abs/2012.15349} {{DynaSent}: A dynamic
  benchmark for sentiment analysis}.
\newblock \emph{arXiv preprint arXiv:2012.15349}.

\bibitem[{Prasad et~al.(2008)Prasad, Dinesh, Lee, Miltsakaki, Robaldo, Joshi,
  and Webber}]{prasad-etal-2008-penn}
Rashmi Prasad, Nikhil Dinesh, Alan Lee, Eleni Miltsakaki, Livio Robaldo,
  Aravind Joshi, and Bonnie Webber. 2008.
\newblock \href
  {http://www.lrec-conf.org/proceedings/lrec2008/pdf/754_paper.pdf} {The {P}enn
  {D}iscourse {T}ree{B}ank 2.0.}
\newblock In \emph{Proceedings of the Sixth International Conference on
  Language Resources and Evaluation ({LREC}'08)}, Marrakech, Morocco. European
  Language Resources Association (ELRA).

\bibitem[{Pruksachatkun et~al.(2020)Pruksachatkun, Yeres, Liu, Phang, Htut,
  Wang, Tenney, and Bowman}]{pruksachatkun-etal-2020-jiant}
Yada Pruksachatkun, Phil Yeres, Haokun Liu, Jason Phang, Phu~Mon Htut, Alex
  Wang, Ian Tenney, and Samuel~R. Bowman. 2020.
\newblock \href {https://doi.org/10.18653/v1/2020.acl-demos.15} {jiant: A
  software toolkit for research on general-purpose text understanding models}.
\newblock In \emph{Proceedings of the 58th Annual Meeting of the Association
  for Computational Linguistics: System Demonstrations}, pages 109--117,
  Online. Association for Computational Linguistics.

\bibitem[{Rahman and Ng(2012)}]{rahman2012resolving}
Altaf Rahman and Vincent Ng. 2012.
\newblock Resolving complex cases of definite pronouns: the winograd schema
  challenge.
\newblock In \emph{Proceedings of the 2012 Joint Conference on Empirical
  Methods in Natural Language Processing and Computational Natural Language
  Learning}, pages 777--789. Association for Computational Linguistics.

\bibitem[{Rajani et~al.(2019)Rajani, McCann, Xiong, and
  Socher}]{rajani2019explain}
Nazneen~Fatema Rajani, Bryan McCann, Caiming Xiong, and Richard Socher. 2019.
\newblock \href {https://arxiv.org/abs/1906.02361} {Explain yourself!
  leveraging language models for commonsense reasoning}.
\newblock In \emph{Proceedings of the 2019 Conference of the Association for
  Computational Linguistics (ACL2019)}.

\bibitem[{Ravichander et~al.(2022)Ravichander, Gardner, and
  Marasovic}]{ravichander-et-al-2022-condaqa}
Abhilasha Ravichander, Matt Gardner, and Ana Marasovic. 2022.
\newblock Condaqa: A contrastive reading comprehension dataset for reasoning
  about negation.

\bibitem[{Ravichander et~al.(2019)Ravichander, Naik, Rose, and
  Hovy}]{ravichander2019equate}
Abhilasha Ravichander, Aakanksha Naik, Carolyn Rose, and Eduard Hovy. 2019.
\newblock Equate: A benchmark evaluation framework for quantitative reasoning
  in natural language inference.
\newblock \emph{arXiv preprint arXiv:1901.03735}.

\bibitem[{Roemmele et~al.(2011)Roemmele, Bejan, and
  Gordon}]{roemmele2011choice}
Melissa Roemmele, Cosmin~Adrian Bejan, and Andrew~S Gordon. 2011.
\newblock Choice of plausible alternatives: An evaluation of commonsense causal
  reasoning.
\newblock In \emph{2011 AAAI Spring Symposium Series}.

\bibitem[{Rogers et~al.(2020)Rogers, Kovaleva, Downey, and
  Rumshisky}]{DBLP:conf/aaai/RogersKDR20}
Anna Rogers, Olga Kovaleva, Matthew Downey, and Anna Rumshisky. 2020.
\newblock \href {https://aaai.org/ojs/index.php/AAAI/article/view/6398}
  {Getting closer to {AI} complete question answering: {A} set of prerequisite
  real tasks}.
\newblock In \emph{The Thirty-Fourth {AAAI} Conference on Artificial
  Intelligence, {AAAI} 2020, The Thirty-Second Innovative Applications of
  Artificial Intelligence Conference, {IAAI} 2020, The Tenth {AAAI} Symposium
  on Educational Advances in Artificial Intelligence, {EAAI} 2020, New York,
  NY, USA, February 7-12, 2020}, pages 8722--8731. {AAAI} Press.

\bibitem[{Rudinger et~al.(2018)Rudinger, Naradowsky, Leonard, and
  V}]{rudinger-EtAl:2018:N18}
Rachel Rudinger, Jason Naradowsky, Brian Leonard, and Benjamin V. 2018.
\newblock Gender bias in coreference resolution.
\newblock In \emph{Proceedings of the 2018 Conference of the North American
  Chapter of the Association for Computational Linguistics: Human Language
  Technologies}, New Orleans, Louisiana. Association for Computational
  Linguistics.

\bibitem[{Sadat and Caragea(2022)}]{sadat-caragea-2022-scinli}
Mobashir Sadat and Cornelia Caragea. 2022.
\newblock \href {https://aclanthology.org/2022.acl-long.511} {{S}ci{NLI}: A
  corpus for natural language inference on scientific text}.
\newblock In \emph{Proceedings of the 60th Annual Meeting of the Association
  for Computational Linguistics (Volume 1: Long Papers)}, pages 7399--7409,
  Dublin, Ireland. Association for Computational Linguistics.

\bibitem[{Saikh et~al.(2022)Saikh, Ghosal, Mittal, Ekbal, and
  Bhattacharyya}]{10.1007/s00799-022-00329-y}
Tanik Saikh, Tirthankar Ghosal, Amish Mittal, Asif Ekbal, and Pushpak
  Bhattacharyya. 2022.
\newblock Scienceqa: A novel resource for question answering on scholarly
  articles.
\newblock \emph{Int. J. Digit. Libr.}

\bibitem[{Schlegel et~al.(2022)Schlegel, Pavlov, and
  Pratt-Hartmann}]{https://doi.org/10.48550/arxiv.2211.05417}
Viktor Schlegel, Kamen~V. Pavlov, and Ian Pratt-Hartmann. 2022.
\newblock \href {https://doi.org/10.48550/ARXIV.2211.05417} {Can transformers
  reason in fragments of natural language?}

\bibitem[{Schler et~al.(2006)Schler, Koppel, Argamon, and
  Pennebaker}]{schler2006effects}
Jonathan Schler, Moshe Koppel, Shlomo Argamon, and James~W Pennebaker. 2006.
\newblock Effects of age and gender on blogging.
\newblock In \emph{AAAI spring symposium: Computational approaches to analyzing
  weblogs}, volume~6, pages 199--205.

\bibitem[{Schuster et~al.(2021)Schuster, Fisch, and
  Barzilay}]{schuster-etal-2021-get}
Tal Schuster, Adam Fisch, and Regina Barzilay. 2021.
\newblock \href {https://doi.org/10.18653/v1/2021.naacl-main.52} {Get your
  vitamin {C}! robust fact verification with contrastive evidence}.
\newblock In \emph{Proceedings of the 2021 Conference of the North American
  Chapter of the Association for Computational Linguistics: Human Language
  Technologies}, pages 624--643, Online. Association for Computational
  Linguistics.

\bibitem[{Sheng and Uthus(2020)}]{sheng2020investigating}
Emily Sheng and David Uthus. 2020.
\newblock \href {http://arxiv.org/abs/2011.02686} {Investigating societal
  biases in a poetry composition system}.

\bibitem[{Shriberg et~al.(2004)Shriberg, Dhillon, Bhagat, Ang, and
  Carvey}]{shriberg2004icsi}
Elizabeth Shriberg, Raj Dhillon, Sonali Bhagat, Jeremy Ang, and Hannah Carvey.
  2004.
\newblock The icsi meeting recorder dialog act (mrda) corpus.
\newblock In \emph{Proceedings of the 5th SIGdial Workshop on Discourse and
  Dialogue at HLT-NAACL 2004}.

\bibitem[{Sileo(2023)}]{sileod22-tasknet}
Damien Sileo. 2023.
\newblock \href {https://doi.org/10.5281/zenodo.7326972} {{tasknet, multitask
  interface between Trainer and datasets}}.

\bibitem[{Sileo and Lernould(2023)}]{sileo2023mindgames}
Damien Sileo and Antoine Lernould. 2023.
\newblock Mindgames: Targeting theory of mind in large language models with
  dynamic epistemic modal logic.
\newblock \emph{arXiv preprint arXiv:2305.03353}.

\bibitem[{Sileo and Moens(2022{\natexlab{a}})}]{sileo-moens-2022-analysis}
Damien Sileo and Marie-Francine Moens. 2022{\natexlab{a}}.
\newblock \href {https://aclanthology.org/2022.lrec-1.67} {Analysis and
  prediction of {NLP} models via task embeddings}.
\newblock In \emph{Proceedings of the Thirteenth Language Resources and
  Evaluation Conference}, pages 633--647, Marseille, France. European Language
  Resources Association.

\bibitem[{Sileo and Moens(2022{\natexlab{b}})}]{sileo2022probing}
Damien Sileo and Marie-Francine Moens. 2022{\natexlab{b}}.
\newblock Probing neural language models for understanding of words of
  estimative probability.
\newblock \emph{arXiv preprint arXiv:2211.03358}.

\bibitem[{Sileo et~al.(2022{\natexlab{a}})Sileo, Muller, Van~de Cruys, and
  Pradel}]{sileo-etal-2022-pragmatics}
Damien Sileo, Philippe Muller, Tim Van~de Cruys, and Camille Pradel.
  2022{\natexlab{a}}.
\newblock \href {https://aclanthology.org/2022.lrec-1.255} {A
  pragmatics-centered evaluation framework for natural language understanding}.
\newblock In \emph{Proceedings of the Thirteenth Language Resources and
  Evaluation Conference}, pages 2382--2394, Marseille, France. European
  Language Resources Association.

\bibitem[{Sileo et~al.(2023)Sileo, Uma, and Moens}]{sileo2023generating}
Damien Sileo, Kanimozhi Uma, and Marie-Francine Moens. 2023.
\newblock Generating multiple-choice questions for medical question answering
  with distractors and cue-masking.
\newblock \emph{arXiv preprint arXiv:2303.07069}.

\bibitem[{Sileo et~al.(2019)Sileo, Van De~Cruys, Pradel, and
  Muller}]{sileo-etal-2019-mining}
Damien Sileo, Tim Van De~Cruys, Camille Pradel, and Philippe Muller. 2019.
\newblock \href {https://www.aclweb.org/anthology/N19-1351} {Mining discourse
  markers for unsupervised sentence representation learning}.
\newblock In \emph{Proceedings of the 2019 Conference of the North {A}merican
  Chapter of the Association for Computational Linguistics: Human Language
  Technologies, Volume 1 (Long and Short Papers)}, pages 3477--3486,
  Minneapolis, Minnesota. Association for Computational Linguistics.

\bibitem[{Sileo et~al.(2022{\natexlab{b}})Sileo, Vossen, and
  Raymaekers}]{sileo-lmrec-2022}
Damien Sileo, Wout Vossen, and Robbe Raymaekers. 2022{\natexlab{b}}.
\newblock Zero-shot recommendation as language modeling.
\newblock In \emph{Advances in Information Retrieval}, pages 223--230, Cham.
  Springer International Publishing.

\bibitem[{Spaeth et~al.(2020)Spaeth, Epstein, Andrew D.~Martin, Ruger, and
  Benesh}]{spaeth2020}
Harold~J. Spaeth, Lee Epstein, Jeffrey A.~Segal Andrew D.~Martin, Theodore~J.
  Ruger, and Sara~C. Benesh. 2020.
\newblock \href {http://Supremecourtdatabase.org} {{Supreme Court Database,
  Version 2020 Release 01}}.
\newblock Washington University Law.

\bibitem[{Srivastava et~al.(2022{\natexlab{a}})Srivastava, Rastogi, Rao, Shoeb,
  Abid, Fisch, Brown, Santoro, Gupta, Garriga-Alonso et~al.}]{bigbench}
Aarohi Srivastava, Abhinav Rastogi, Abhishek Rao, Abu Awal~Md Shoeb, Abubakar
  Abid, Adam Fisch, Adam~R Brown, Adam Santoro, Aditya Gupta, Adri{\`a}
  Garriga-Alonso, et~al. 2022{\natexlab{a}}.
\newblock Beyond the imitation game: Quantifying and extrapolating the
  capabilities of language models.
\newblock \emph{arXiv preprint arXiv:2206.04615}.

\bibitem[{Srivastava et~al.(2022{\natexlab{b}})Srivastava, Rastogi, Rao, Shoeb,
  Abid, Fisch, Brown, Santoro, Gupta, Garriga-Alonso
  et~al.}]{srivastava2022beyond}
Aarohi Srivastava, Abhinav Rastogi, Abhishek Rao, Abu Awal~Md Shoeb, Abubakar
  Abid, Adam Fisch, Adam~R Brown, Adam Santoro, Aditya Gupta, Adria
  Garriga-Alonso, et~al. 2022{\natexlab{b}}.
\newblock Beyond the imitation game: Quantifying and extrapolating the
  capabilities of language models.
\newblock \emph{arXiv preprint arXiv:2206.04615}.

\bibitem[{Stiennon et~al.(2020)Stiennon, Ouyang, Wu, Ziegler, Lowe, Voss,
  Radford, Amodei, and Christiano}]{stienon2020learning}
Nisan Stiennon, Long Ouyang, Jeff Wu, Daniel~M. Ziegler, Ryan Lowe, Chelsea
  Voss, Alec Radford, Dario Amodei, and Paul Christiano. 2020.
\newblock Learning to summarize from human feedback.
\newblock In \emph{NeurIPS}.

\bibitem[{Sun et~al.(2019)Sun, Yu, Chen, Yu, Choi, and Cardie}]{sundream2018}
Kai Sun, Dian Yu, Jianshu Chen, Dong Yu, Yejin Choi, and Claire Cardie. 2019.
\newblock \href {https://arxiv.org/abs/1902.00164v1} {{DREAM}: A challenge
  dataset and models for dialogue-based reading comprehension}.
\newblock \emph{Transactions of the Association for Computational Linguistics}.

\bibitem[{Szomiu and Groza(2021)}]{szomiu2021puzzle}
Roxana Szomiu and Adrian Groza. 2021.
\newblock A puzzle-based dataset for natural language inference.
\newblock \emph{arXiv preprint arXiv:2112.05742}.

\bibitem[{Tafjord et~al.("2019")Tafjord, Gardner, Lin, and Clark}]{quartz}
Oyvind Tafjord, Matt Gardner, Kevin Lin, and Peter Clark. "2019".
\newblock "quartz: An open-domain dataset of qualitative relationship
  questions".

\bibitem[{Talmor et~al.(2019)Talmor, Herzig, Lourie, and
  Berant}]{talmor-etal-2019-commonsenseqa}
Alon Talmor, Jonathan Herzig, Nicholas Lourie, and Jonathan Berant. 2019.
\newblock \href {https://doi.org/10.18653/v1/N19-1421} {{C}ommonsense{QA}: A
  question answering challenge targeting commonsense knowledge}.
\newblock In \emph{Proceedings of the 2019 Conference of the North {A}merican
  Chapter of the Association for Computational Linguistics: Human Language
  Technologies, Volume 1 (Long and Short Papers)}, pages 4149--4158,
  Minneapolis, Minnesota. Association for Computational Linguistics.

\bibitem[{Tandon et~al.(2019)Tandon, Mishra, Sakaguchi, Bosselut, and
  Clark}]{wiqa}
Niket Tandon, Bhavana~Dalvi Mishra, Keisuke Sakaguchi, Antoine Bosselut, and
  Peter Clark. 2019.
\newblock Wiqa: A dataset for "what if..." reasoning over procedural text.
\newblock \emph{arXiv:1909.04739v1}.

\bibitem[{Tarunesh et~al.(2021)Tarunesh, Aditya, and
  Choudhury}]{Tarunesh2021TrustingRO}
Ishan Tarunesh, Somak Aditya, and Monojit Choudhury. 2021.
\newblock Trusting roberta over bert: Insights from checklisting the natural
  language inference task.
\newblock \emph{ArXiv}, abs/2107.07229.

\bibitem[{Thompson et~al.(1993)Thompson, Anderson, Bard, Doherty-Sneddon,
  Newlands, and Sotillo}]{thompson1993hcrc}
Henry~S Thompson, Anne~H Anderson, Ellen~Gurman Bard, Gwyneth Doherty-Sneddon,
  Alison Newlands, and Cathy Sotillo. 1993.
\newblock The hcrc map task corpus: natural dialogue for speech recognition.
\newblock In \emph{HUMAN LANGUAGE TECHNOLOGY: Proceedings of a Workshop Held at
  Plainsboro, New Jersey, March 21-24, 1993}.

\bibitem[{Thukral et~al.(2021)Thukral, Kukreja, and
  Kavouras}]{thukral-etal-2021-probing}
Shivin Thukral, Kunal Kukreja, and Christian Kavouras. 2021.
\newblock \href {https://doi.org/10.18653/v1/2021.blackboxnlp-1.31} {Probing
  language models for understanding of temporal expressions}.
\newblock In \emph{Proceedings of the Fourth BlackboxNLP Workshop on Analyzing
  and Interpreting Neural Networks for NLP}, pages 396--406, Punta Cana,
  Dominican Republic. Association for Computational Linguistics.

\bibitem[{Tjong Kim~Sang and
  De~Meulder(2003)}]{tjong-kim-sang-de-meulder-2003-introduction}
Erik~F. Tjong Kim~Sang and Fien De~Meulder. 2003.
\newblock \href {https://www.aclweb.org/anthology/W03-0419} {Introduction to
  the {C}o{NLL}-2003 shared task: Language-independent named entity
  recognition}.
\newblock In \emph{Proceedings of the Seventh Conference on Natural Language
  Learning at {HLT}-{NAACL} 2003}, pages 142--147.

\bibitem[{Tuggener et~al.(2020)Tuggener, von Daniken, Peetz, and
  Cieliebak}]{tuggener-etal-2020-ledgar}
Don Tuggener, Pius von Daniken, Thomas Peetz, and Mark Cieliebak. 2020.
\newblock \href {https://aclanthology.org/2020.lrec-1.155} {{LEDGAR}: A
  large-scale multi-label corpus for text classification of legal provisions in
  contracts}.
\newblock In \emph{Proceedings of the 12th Language Resources and Evaluation
  Conference}, Marseille, France.

\bibitem[{Tunstall et~al.(2022)Tunstall, Reimers, Jo, Bates, Korat, Wasserblat,
  and Pereg}]{tunstall2022efficient}
Lewis Tunstall, Nils Reimers, Unso Eun~Seo Jo, Luke Bates, Daniel Korat, Moshe
  Wasserblat, and Oren Pereg. 2022.
\newblock Efficient few-shot learning without prompts.
\newblock \emph{arXiv preprint arXiv:2209.11055}.

\bibitem[{Van~Pelt and Sorokin(2012)}]{van2012designing}
Chris Van~Pelt and Alex Sorokin. 2012.
\newblock Designing a scalable crowdsourcing platform.
\newblock In \emph{Proceedings of the 2012 ACM SIGMOD International Conference
  on Management of Data}, pages 765--766.

\bibitem[{Veyseh et~al.(2020)Veyseh, Dernoncourt, Tran, and
  Nguyen}]{veyseh-et-al-2020-what}
Amir Pouran~Ben Veyseh, Franck Dernoncourt, Quan~Hung Tran, and Thien~Huu
  Nguyen. 2020.
\newblock {What Does This Acronym Mean? Introducing a New Dataset for Acronym
  Identification and Disambiguation}.
\newblock In \emph{Proceedings of COLING}.

\bibitem[{Vidgen et~al.(2021)Vidgen, Thrush, Waseem, and
  Kiela}]{vidgen2021learning}
Bertie Vidgen, Tristan Thrush, Zeerak Waseem, and Douwe Kiela. 2021.
\newblock Learning from the worst: Dynamically generated datasets to improve
  online hate detection.
\newblock In \emph{Proceedings of the 59th Annual Meeting of the Association
  for Computational Linguistics and the 11th International Joint Conference on
  Natural Language Processing (Volume 1: Long Papers)}, pages 1667--1682.

\bibitem[{Vilares and
  Gomez-Rodriguez(2019)}]{vilares-gomez-rodriguez-2019-head}
David Vilares and Carlos Gomez-Rodriguez. 2019.
\newblock \href {https://doi.org/10.18653/v1/P19-1092} {{HEAD}-{QA}: A
  healthcare dataset for complex reasoning}.
\newblock In \emph{Proceedings of the 57th Annual Meeting of the Association
  for Computational Linguistics}, pages 960--966, Florence, Italy. Association
  for Computational Linguistics.

\bibitem[{Wallace et~al.(2022)Wallace, Williams, Jia, and
  Kiela}]{Wallace2022Dynamic}
Eric Wallace, Adina Williams, Robin Jia, and Douwe Kiela. 2022.
\newblock Analyzing dynamic adversarial training data in the limit.
\newblock In \emph{Findings of the Association for Computational Linguistics}.

\bibitem[{Wang et~al.(2019{\natexlab{a}})Wang, Singh, Michael, Hill, Levy, and
  Bowman}]{wang2019glue}
Alex Wang, Amanpreet Singh, Julian Michael, Felix Hill, Omer Levy, and
  Samuel~R. Bowman. 2019{\natexlab{a}}.
\newblock {GLUE}: A multi-task benchmark and analysis platform for natural
  language understanding.
\newblock In the Proceedings of ICLR.

\bibitem[{Wang et~al.(2019{\natexlab{b}})Wang, He, and
  Zhou}]{wang-etal-2019-spherere}
Chengyu Wang, Xiaofeng He, and Aoying Zhou. 2019{\natexlab{b}}.
\newblock \href {https://doi.org/10.18653/v1/P19-1169} {{S}phere{RE}:
  Distinguishing lexical relations with hyperspherical relation embeddings}.
\newblock In \emph{Proceedings of the 57th Annual Meeting of the Association
  for Computational Linguistics}, pages 1727--1737, Florence, Italy.
  Association for Computational Linguistics.

\bibitem[{Wang(2017)}]{wang-2017-liar}
William~Yang Wang. 2017.
\newblock \href {https://doi.org/10.18653/v1/P17-2067} {{``}liar, liar pants on
  fire{''}: A new benchmark dataset for fake news detection}.
\newblock In \emph{Proceedings of the 55th Annual Meeting of the Association
  for Computational Linguistics (Volume 2: Short Papers)}, pages 422--426,
  Vancouver, Canada. Association for Computational Linguistics.

\bibitem[{Wang et~al.(2022)Wang, Mishra, Alipoormolabashi, Kordi, Mirzaei,
  Arunkumar, Ashok, Dhanasekaran, Naik, Stap et~al.}]{supernaturalinstructions}
Yizhong Wang, Swaroop Mishra, Pegah Alipoormolabashi, Yeganeh Kordi, Amirreza
  Mirzaei, Anjana Arunkumar, Arjun Ashok, Arut~Selvan Dhanasekaran, Atharva
  Naik, David Stap, et~al. 2022.
\newblock Super-naturalinstructions:generalization via declarative instructions
  on 1600+ tasks.
\newblock In \emph{EMNLP}.

\bibitem[{Warstadt et~al.(2019)Warstadt, Parrish, Liu, Mohananey, Peng, Wang,
  and Bowman}]{warstadt2019blimp}
Alex Warstadt, Alicia Parrish, Haokun Liu, Anhad Mohananey, Wei Peng, Sheng-Fu
  Wang, and Samuel~R Bowman. 2019.
\newblock Blimp: A benchmark of linguistic minimal pairs for english.
\newblock \emph{arXiv preprint arXiv:1912.00582}.

\bibitem[{Warstadt et~al.(2018)Warstadt, Singh, and
  Bowman}]{warstadt2018neural}
Alex Warstadt, Amanpreet Singh, and Samuel~R Bowman. 2018.
\newblock Neural network acceptability judgments.
\newblock \emph{arXiv preprint arXiv:1805.12471}.

\bibitem[{Welbl et~al.(2018)Welbl, Stenetorp, and
  Riedel}]{welbl2018constructing}
Johannes Welbl, Pontus Stenetorp, and Sebastian Riedel. 2018.
\newblock \href {http://arxiv.org/abs/1710.06481} {Constructing datasets for
  multi-hop reading comprehension across documents}.

\bibitem[{Weston et~al.(2015)Weston, Bordes, Chopra, Rush, Van~Merrienboer,
  Joulin, and Mikolov}]{weston2015towards}
Jason Weston, Antoine Bordes, Sumit Chopra, Alexander~M Rush, Bart
  Van~Merrienboer, Armand Joulin, and Tomas Mikolov. 2015.
\newblock Towards ai-complete question answering: A set of prerequisite toy
  tasks.
\newblock \emph{arXiv preprint arXiv:1502.05698}.

\bibitem[{Williams et~al.(2018)Williams, Nangia, and Bowman}]{N18-1101}
Adina Williams, Nikita Nangia, and Samuel Bowman. 2018.
\newblock \href {http://aclweb.org/anthology/N18-1101} {A broad-coverage
  challenge corpus for sentence understanding through inference}.
\newblock In \emph{Proceedings of NAACL2018}, pages 1112--1122. Association for
  Computational Linguistics.

\bibitem[{Wolf et~al.(2020)Wolf, Lhoest, von Platen, Jernite, Drame, Plu,
  Chaumond, Delangue, Ma, Thakur, Patil, Davison, Scao, Sanh, Xu, Patry,
  McMillan-Major, Brandeis, Gugger, Lagunas, Debut, Funtowicz, Moi, Rush,
  Schmidd, Cistac, Muštar, Boudier, and Tordjmann}]{2020HuggingFace-datasets}
Thomas Wolf, Quentin Lhoest, Patrick von Platen, Yacine Jernite, Mariama Drame,
  Julien Plu, Julien Chaumond, Clement Delangue, Clara Ma, Abhishek Thakur,
  Suraj Patil, Joe Davison, Teven~Le Scao, Victor Sanh, Canwen Xu, Nicolas
  Patry, Angie McMillan-Major, Simon Brandeis, Sylvain Gugger, François
  Lagunas, Lysandre Debut, Morgan Funtowicz, Anthony Moi, Sasha Rush, Philipp
  Schmidd, Pierric Cistac, Victor Muštar, Jeff Boudier, and Anna Tordjmann.
  2020.
\newblock Datasets.
\newblock \emph{GitHub. Note: https://github.com/huggingface/datasets}, 1.

\bibitem[{Wright and Augenstein(2021)}]{wright2021exaggeration}
Dustin Wright and Isabelle Augenstein. 2021.
\newblock {Semi-Supervised Exaggeration Detection of Health Science Press
  Releases}.
\newblock In \emph{Proceedings of EMNLP}. Association for Computational
  Linguistics.

\bibitem[{Y et~al.(2015)Y, W, and Meek}]{YangYihMeek:EMNLP2015:WikiQA}
Yang Y, Yih W, and C~Meek. 2015.
\newblock \href {https://doi.org/10.18653/v1/D15-1237} {{WikiQA: A Challenge
  Dataset for Open-Domain Question Answering}}.
\newblock page 2013–2018.

\bibitem[{Yanaka et~al.(2019{\natexlab{a}})Yanaka, Mineshima, Bekki, Inui,
  Sekine, Abzianidze, and Bos}]{yanaka-etal-2019-neural}
Hitomi Yanaka, Koji Mineshima, Daisuke Bekki, Kentaro Inui, Satoshi Sekine,
  Lasha Abzianidze, and Johan Bos. 2019{\natexlab{a}}.
\newblock Can neural networks understand monotonicity reasoning?
\newblock In \emph{Proceedings of the 2019 ACL Workshop BlackboxNLP: Analyzing
  and Interpreting Neural Networks for NLP}, pages 31--40.

\bibitem[{Yanaka et~al.(2019{\natexlab{b}})Yanaka, Mineshima, Bekki, Inui,
  Sekine, Abzianidze, and Bos}]{yanaka-EtAl:2019:starsem}
Hitomi Yanaka, Koji Mineshima, Daisuke Bekki, Kentaro Inui, Satoshi Sekine,
  Lasha Abzianidze, and Johan Bos. 2019{\natexlab{b}}.
\newblock Help: A dataset for identifying shortcomings of neural models in
  monotonicity reasoning.
\newblock In \emph{Proceedings of the Eighth Joint Conference on Lexical and
  Computational Semantics (*SEM2019)}.

\bibitem[{Yanaka et~al.(2021)Yanaka, Mineshima, and
  Inui}]{yanaka-etal-2021-exploring}
Hitomi Yanaka, Koji Mineshima, and Kentaro Inui. 2021.
\newblock Exploring transitivity in neural {NLI} models through veridicality.
\newblock In \emph{Proceedings of the 16th Conference of the European Chapter
  of the Association for Computational Linguistics: Main Volume}, pages
  920--934.

\bibitem[{Ye et~al.(2021)Ye, Lin, and Ren}]{ye-etal-2021-crossfit}
Qinyuan Ye, Bill~Yuchen Lin, and Xiang Ren. 2021.
\newblock \href {https://doi.org/10.18653/v1/2021.emnlp-main.572}
  {{C}ross{F}it: A few-shot learning challenge for cross-task generalization in
  {NLP}}.
\newblock In \emph{Proceedings of the 2021 Conference on Empirical Methods in
  Natural Language Processing}, pages 7163--7189, Online and Punta Cana,
  Dominican Republic. Association for Computational Linguistics.

\bibitem[{Yu et~al.(2020{\natexlab{a}})Yu, Kumar, Gupta, Levine, Hausman, and
  Finn}]{yu2020gradient}
Tianhe Yu, Saurabh Kumar, Abhishek Gupta, Sergey Levine, Karol Hausman, and
  Chelsea Finn. 2020{\natexlab{a}}.
\newblock Gradient surgery for multi-task learning.
\newblock \emph{Advances in Neural Information Processing Systems},
  33:5824--5836.

\bibitem[{Yu et~al.(2020{\natexlab{b}})Yu, Jiang, Dong, and
  Feng}]{yu2020reclor}
Weihao Yu, Zihang Jiang, Yanfei Dong, and Jiashi Feng. 2020{\natexlab{b}}.
\newblock Reclor: A reading comprehension dataset requiring logical reasoning.
\newblock In \emph{International Conference on Learning Representations
  (ICLR)}.

\bibitem[{Zeldes(2017)}]{Zeldes2017}
Amir Zeldes. 2017.
\newblock \href {https://doi.org/http://dx.doi.org/10.1007/s10579-016-9343-x}
  {The {GUM} corpus: Creating multilayer resources in the classroom}.
\newblock \emph{Language Resources and Evaluation}, 51(3):581--612.

\bibitem[{Zellers et~al.(2018)Zellers, Bisk, Schwartz, and
  Choi}]{zellers2018swagaf}
Rowan Zellers, Yonatan Bisk, Roy Schwartz, and Yejin Choi. 2018.
\newblock Swag: A large-scale adversarial dataset for grounded commonsense
  inference.
\newblock In \emph{Proceedings of the 2018 Conference on Empirical Methods in
  Natural Language Processing (EMNLP)}.

\bibitem[{Zellers et~al.(2019)Zellers, Holtzman, Bisk, Farhadi, and
  Choi}]{zellers2019hellaswag}
Rowan Zellers, Ari Holtzman, Yonatan Bisk, Ali Farhadi, and Yejin Choi. 2019.
\newblock Hellaswag: Can a machine really finish your sentence?
\newblock In \emph{Proceedings of the 57th Annual Meeting of the Association
  for Computational Linguistics}.

\bibitem[{Zeman et~al.(2020)Zeman, Nivre, Abrams, Ackermann, Aepli, Aghaei,
  Agic, Ahmadi, Ahrenberg, Ajede, Aleksandraviviute, Alfina, Antonsen,
  Aplonova, Aquino, Aragon, Aranzabe, Arnardottir, Arutie, Arwidarasti,
  Asahara, Ateyah, Atmaca, Attia, Atutxa, Augustinus, Badmaeva, Balasubramani,
  Ballesteros, Banerjee, Bank, Barbu~Mititelu, Basmov, Batchelor, Bauer, Bedir,
  Bengoetxea, Berk, Berzak, Bhat, Bhat, Biagetti, Bick, Bielinskiene,
  Bjarnadottir, Blokland, Bobicev, Boizou, Borges~Volker, Borstell, Bosco,
  Bouma, Bowman, Boyd, Brokaite, Burchardt, Candito, Caron, Caron, Cavalcanti,
  Cebiroglu~Eryigit, Cecchini, Celano, Ceplo, Cetin, Cetinoglu, Chalub, Chi,
  Cho, Choi, Chun, Cignarella, Cinkova, Collomb, Coltekin, Connor, Courtin,
  Davidson, de~Marneffe, de~Paiva, Derin, de~Souza, Diaz~de Ilarraza,
  Dickerson, Dinakaramani, Dione, Dirix, Dobrovoljc, Dozat, Droganova, Dwivedi,
  Eckhoff, Eli, Elkahky, Ephrem, Erina, Erjavec, Etienne, Evelyn, Facundes,
  Farkas, Fernanda, Fernandez~Alcalde, Foster, Freitas, Fujita, Gajdosova,
  Galbraith, Garcia, Gardenfors, Garza, Gerardi, Gerdes, Ginter, Goenaga,
  Gojenola, Gokirmak, Goldberg, Gomez~Guinovart, Gonzalez~Saavedra, Griciute,
  Grioni, Grobol, Gruzitis, Guillaume, Guillot-Barbance, Gungor, Habash,
  Hafsteinsson, Hajiv, Hajiv~jr., Hamalainen, Ha~My, Han, Hanifmuti, Hardwick,
  Harris, Haug, Heinecke, Hellwig, Hennig, Hladka, Hlavavova, Hociung, Hohle,
  Huber, Hwang, Ikeda, Ingason, Ion, Irimia, Ishola, Jelinek, Johannsen,
  Jonsdottir, Jorgensen, Juutinen, K, Kacikara, Kaasen, Kabaeva, Kahane,
  Kanayama, Kanerva, Katz, Kayadelen, Kenney, Kettnerova, Kirchner,
  Klementieva, Kohn, Koksal, Kopacewicz, Korkiakangas, Kotsyba, Kovalevskaite,
  Krek, Krishnamurthy, Kwak, Laippala, Lam, Lambertino, Lando, Larasati,
  Lavrentiev, Lee, Le~Hong, Lenci, Lertpradit, Leung, Levina, Li, Li, Li, Li,
  Lim, Linden, Ljubesic, Loginova, Luthfi, Luukko, Lyashevskaya, Lynn,
  Macketanz, Makazhanov, Mandl, Manning, Manurung, Maranduc, Marcek,
  Marheinecke, Martinez~Alonso, Martins, Masek, Matsuda, Matsumoto, M, M,
  Mendonca, Miekka, Mischenkova, Misirpashayeva, Missila, Mititelu, Mitrofan,
  Miyao, Mojiri~Foroushani, Moloodi, Montemagni, More, Moreno~Romero, Mori,
  Mori, Morioka, Moro, Mortensen, Moskalevskyi, Muischnek, Munro, Murawaki,
  Muurisep, Nainwani, Nakhle, Navarro~Horniacek, Nedoluzhko, Nevpore-Berzkalne,
  Nguyen~Thd, Nguyen Thd~Minh, Nikaido, Nikolaev, Nitisaroj, Nourian, Nurmi,
  Ojala, Ojha, Oluokun, Omura, Onwuegbuzia, Osenova, Ostling, Ovrelid, Ozatec,
  Ozgur, Ozturk~Bacaran, Partanen, Pascual, Passarotti, Patejuk,
  Paulino-Passos, Peljak-Lapinska, Peng, Perez, Perkova, Perrier, Petrov,
  Petrova, Phelan, Piitulainen, Pirinen, Pitler, Plank, Poibeau, Ponomareva,
  Popel, Pretkalnina, Prevost, Prokopidis, Przepiorkowski, Puolakainen,
  Pyysalo, Qi, Raabis, Rademaker, Rama, Ramasamy, Ramisch, Rashel, Rasooli,
  Ravishankar, Real, Rebeja, Reddy, Rehm, Riabov, Riesler, Rimkute, Rinaldi,
  Rituma, Rocha, Rognvaldsson, Romanenko, Rosa, Roca, Rovati, Rudina, Rueter,
  Runarsson, Sadde, Safari, Sagot, Sahala, Saleh, Salomoni, Samardzic, Samson,
  Sanguinetti, Sarg, Saulite, Sawanakunanon, Scannell, Scarlata, Schneider,
  Schuster, Seddah, Seeker, Seraji, Shen, Shimada, Shirasu, Shohibussirri,
  Sichinava, Sigursson, Silveira, Silveira, Simi, Simionescu, Simko, vimkova,
  Simov, Skachedubova, Smith, Soares-Bastos, Spadine, Steingrimsson, Stella,
  Straka, Strickland, Strnadova, Suhr, Sulestio, Sulubacak, Suzuki, Szanto,
  Taji, Takahashi, Tamburini, Tan, Tanaka, Tella, Tellier, Thomas, Torga,
  Toska, Trosterud, Trukhina, Tsarfaty, Turk, Tyers, Uematsu, Untilov, Uresova,
  Uria, Uszkoreit, Utka, Vajjala, van Niekerk, van Noord, Varga, Villemonte
  de~la Clergerie, Vincze, Wakasa, Wallenberg, Wallin, Walsh, Wang, Washington,
  Wendt, Widmer, Williams, Wiren, Wittern, Woldemariam, Wong, Wroblewska, Yako,
  Yamashita, Yamazaki, Yan, Yasuoka, Yavrumyan, Yu, Zabokrtsky, Zahra, Zeldes,
  Zhu, and Zhuravleva}]{11234/1-3424}
Daniel Zeman, Joakim Nivre, Mitchell Abrams, Elia Ackermann, Noemi Aepli, Hamid
  Aghaei, veljko Agic, Amir Ahmadi, Lars Ahrenberg, Chika~Kennedy Ajede,
  Gabriele Aleksandraviviute, Ika Alfina, Lene Antonsen, Katya Aplonova,
  Angelina Aquino, Carolina Aragon, Maria~Jesus Aranzabe, torunn Arnardottir,
  Gashaw Arutie, Jessica~Naraiswari Arwidarasti, Masayuki Asahara, Luma Ateyah,
  Furkan Atmaca, Mohammed Attia, Aitziber Atutxa, Liesbeth Augustinus, Elena
  Badmaeva, Keerthana Balasubramani, Miguel Ballesteros, Esha Banerjee,
  Sebastian Bank, Verginica Barbu~Mititelu, Victoria Basmov, Colin Batchelor,
  John Bauer, Seyyit~Talha Bedir, Kepa Bengoetxea, Gozde Berk, Yevgeni Berzak,
  Irshad~Ahmad Bhat, Riyaz~Ahmad Bhat, Erica Biagetti, Eckhard Bick, Agne
  Bielinskiene, Kristin Bjarnadottir, Rogier Blokland, Victoria Bobicev, Loic
  Boizou, Emanuel Borges~Volker, Carl Borstell, Cristina Bosco, Gosse Bouma,
  Sam Bowman, Adriane Boyd, Kristina Brokaite, Aljoscha Burchardt, Marie
  Candito, Bernard Caron, Gauthier Caron, Tatiana Cavalcanti, Gulsen
  Cebiroglu~Eryigit, Flavio~Massimiliano Cecchini, Giuseppe G.~A. Celano,
  Slavomir Ceplo, Savas Cetin, Ozlem Cetinoglu, Fabricio Chalub, Ethan Chi,
  Yongseok Cho, Jinho Choi, Jayeol Chun, Alessandra~T. Cignarella, Silvie
  Cinkova, Aurelie Collomb, Cagri Coltekin, Miriam Connor, Marine Courtin,
  Elizabeth Davidson, Marie-Catherine de~Marneffe, Valeria de~Paiva,
  Mehmet~Oguz Derin, Elvis de~Souza, Arantza Diaz~de Ilarraza, Carly Dickerson,
  Arawinda Dinakaramani, Bamba Dione, Peter Dirix, Kaja Dobrovoljc, Timothy
  Dozat, Kira Droganova, Puneet Dwivedi, Hanne Eckhoff, Marhaba Eli, Ali
  Elkahky, Binyam Ephrem, Olga Erina, Tomaz Erjavec, Aline Etienne, Wograine
  Evelyn, Sidney Facundes, Richard Farkas, Marilia Fernanda, Hector
  Fernandez~Alcalde, Jennifer Foster, Claudia Freitas, Kazunori Fujita,
  Katarina Gajdosova, Daniel Galbraith, Marcos Garcia, Moa Gardenfors,
  Sebastian Garza, Fabricio~Ferraz Gerardi, Kim Gerdes, Filip Ginter, Iakes
  Goenaga, Koldo Gojenola, Memduh Gokirmak, Yoav Goldberg, Xavier
  Gomez~Guinovart, Berta Gonzalez~Saavedra, Bernadeta Griciute, Matias Grioni,
  Loic Grobol, Normunds Gruzitis, Bruno Guillaume, Celine Guillot-Barbance,
  Tunga Gungor, Nizar Habash, Hinrik Hafsteinsson, Jan Hajiv, Jan Hajiv~jr.,
  Mika Hamalainen, Linh Ha~My, Na-Rae Han, Muhammad~Yudistira Hanifmuti, Sam
  Hardwick, Kim Harris, Dag Haug, Johannes Heinecke, Oliver Hellwig, Felix
  Hennig, Barbora Hladka, Jaroslava Hlavavova, Florinel Hociung, Petter Hohle,
  Eva Huber, Jena Hwang, Takumi Ikeda, Anton~Karl Ingason, Radu Ion, Elena
  Irimia, dlajide Ishola, Tomav Jelinek, Anders Johannsen, Hildur Jonsdottir,
  Fredrik Jorgensen, Markus Juutinen, Sarveswaran K, Huner Kacikara, Andre
  Kaasen, Nadezhda Kabaeva, Sylvain Kahane, Hiroshi Kanayama, Jenna Kanerva,
  Boris Katz, Tolga Kayadelen, Jessica Kenney, Vaclava Kettnerova, Jesse
  Kirchner, Elena Klementieva, Arne Kohn, Abdullatif Koksal, Kamil Kopacewicz,
  Timo Korkiakangas, Natalia Kotsyba, Jolanta Kovalevskaite, Simon Krek,
  Parameswari Krishnamurthy, Sookyoung Kwak, Veronika Laippala, Lucia Lam,
  Lorenzo Lambertino, Tatiana Lando, Septina~Dian Larasati, Alexei Lavrentiev,
  John Lee, Phng Le~Hong, Alessandro Lenci, Saran Lertpradit, Herman Leung,
  Maria Levina, Cheuk~Ying Li, Josie Li, Keying Li, Yuan Li, K~Lim, Krister
  Linden, Nikola Ljubesic, Olga Loginova, Andry Luthfi, Mikko Luukko, Olga
  Lyashevskaya, Teresa Lynn, Vivien Macketanz, Aibek Makazhanov, Michael Mandl,
  Christopher Manning, Ruli Manurung, Catalina Maranduc, David Marcek, Katrin
  Marheinecke, Hector Martinez~Alonso, Andre Martins, Jan Masek, Hiroshi
  Matsuda, Yuji Matsumoto, Ryan M, Sarah M, Gustavo Mendonca, Niko Miekka,
  Karina Mischenkova, Margarita Misirpashayeva, Anna Missila, Catalin Mititelu,
  Maria Mitrofan, Yusuke Miyao, A~Mojiri~Foroushani, Amirsaeid Moloodi,
  Simonetta Montemagni, Amir More, Laura Moreno~Romero, Keiko~Sophie Mori,
  Shinsuke Mori, Tomohiko Morioka, Shigeki Moro, Bjartur Mortensen, Bohdan
  Moskalevskyi, Kadri Muischnek, Robert Munro, Yugo Murawaki, Kaili Muurisep,
  Pinkey Nainwani, Mariam Nakhle, Juan~Ignacio Navarro~Horniacek, Anna
  Nedoluzhko, Gunta Nevpore-Berzkalne, Lng Nguyen~Thd, Huyen Nguyen Thd~Minh,
  Yoshihiro Nikaido, Vitaly Nikolaev, Rattima Nitisaroj, Alireza Nourian, Hanna
  Nurmi, Stina Ojala, Atul~Kr. Ojha, Adedayd Oluokun, Mai Omura, Emeka
  Onwuegbuzia, Petya Osenova, Robert Ostling, Lilja Ovrelid, caziye~Betul
  Ozatec, Arzucan Ozgur, Balkiz Ozturk~Bacaran, Niko Partanen, Elena Pascual,
  Marco Passarotti, Agnieszka Patejuk, Guilherme Paulino-Passos, Angelika
  Peljak-Lapinska, Siyao Peng, Cenel-Augusto Perez, Natalia Perkova, Guy
  Perrier, Slav Petrov, Daria Petrova, Jason Phelan, Jussi Piitulainen, Tommi~A
  Pirinen, Emily Pitler, Barbara Plank, Thierry Poibeau, Larisa Ponomareva,
  Martin Popel, Lauma Pretkalnina, Sophie Prevost, Prokopis Prokopidis, Adam
  Przepiorkowski, Tiina Puolakainen, Sampo Pyysalo, Peng Qi, Andriela Raabis,
  Alexandre Rademaker, Taraka Rama, Loganathan Ramasamy, Carlos Ramisch, Fam
  Rashel, Mohammad~Sadegh Rasooli, Vinit Ravishankar, Livy Real, Petru Rebeja,
  Siva Reddy, Georg Rehm, Ivan Riabov, Michael Riesler, Erika Rimkute, Larissa
  Rinaldi, Laura Rituma, Luisa Rocha, Eirikur Rognvaldsson, Mykhailo Romanenko,
  Rudolf Rosa, Valentin Roca, Davide Rovati, Olga Rudina, Jack Rueter, Kristjan
  Runarsson, Shoval Sadde, Pegah Safari, Benoit Sagot, Aleksi Sahala, Shadi
  Saleh, Alessio Salomoni, Tanja Samardzic, Stephanie Samson, Manuela
  Sanguinetti, Dage Sarg, Baiba Saulite, Yanin Sawanakunanon, Kevin Scannell,
  Salvatore Scarlata, Nathan Schneider, Sebastian Schuster, Djame Seddah,
  Wolfgang Seeker, Mojgan Seraji, Mo~Shen, Atsuko Shimada, Hiroyuki Shirasu,
  Muh Shohibussirri, Dmitry Sichinava, Einar~Freyr Sigursson, Aline Silveira,
  Natalia Silveira, Maria Simi, Radu Simionescu, Katalin Simko, Maria vimkova,
  Kiril Simov, Maria Skachedubova, Aaron Smith, Isabela Soares-Bastos, Carolyn
  Spadine, Steintor Steingrimsson, Antonio Stella, Milan Straka, Emmett
  Strickland, Jana Strnadova, Alane Suhr, Yogi~Lesmana Sulestio, Umut
  Sulubacak, Shingo Suzuki, Zsolt Szanto, Dima Taji, Yuta Takahashi, Fabio
  Tamburini, Mary Ann~C. Tan, Takaaki Tanaka, Samson Tella, Isabelle Tellier,
  Guillaume Thomas, Liisi Torga, Marsida Toska, Trond Trosterud, Anna Trukhina,
  Reut Tsarfaty, Utku Turk, Francis Tyers, Sumire Uematsu, Roman Untilov,
  Zdenka Uresova, Larraitz Uria, Hans Uszkoreit, Andrius Utka, Sowmya Vajjala,
  Daniel van Niekerk, Gertjan van Noord, Viktor Varga, Eric Villemonte de~la
  Clergerie, Veronika Vincze, Aya Wakasa, Joel~C. Wallenberg, Lars Wallin,
  Abigail Walsh, Jing~Xian Wang, Jonathan~North Washington, Maximilan Wendt,
  Paul Widmer, Seyi Williams, Mats Wiren, Christian Wittern, Tsegay
  Woldemariam, Tak-sum Wong, Alina Wroblewska, Mary Yako, Kayo Yamashita, Naoki
  Yamazaki, Chunxiao Yan, Koichi Yasuoka, Marat~M. Yavrumyan, Zhuoran Yu,
  Zdenek Zabokrtsky, Shorouq Zahra, Amir Zeldes, Hanzhi Zhu, and Anna
  Zhuravleva. 2020.
\newblock \href {http://hdl.handle.net/11234/1-3424} {Universal dependencies
  2.7}.
\newblock {LINDAT}/{CLARIAH}-{CZ} digital library at the Institute of Formal
  and Applied Linguistics ({{\'U}FAL}), Faculty of Mathematics and Physics,
  Charles University.

\bibitem[{Zhang et~al.(2015{\natexlab{a}})Zhang, Zhao, and
  LeCun}]{zhang2015character}
Xiang Zhang, Junbo Zhao, and Yann LeCun. 2015{\natexlab{a}}.
\newblock Character-level convolutional networks for text classification.
\newblock In \emph{Advances in neural information processing systems}, pages
  649--657.

\bibitem[{Zhang et~al.(2015{\natexlab{b}})Zhang, Zhao, and
  LeCun}]{Zhang2015CharacterlevelCN}
Xiang Zhang, Junbo~Jake Zhao, and Yann LeCun. 2015{\natexlab{b}}.
\newblock Character-level convolutional networks for text classification.
\newblock In \emph{NIPS}.

\bibitem[{Zhang et~al.(2019)Zhang, Baldridge, and He}]{paws2019naacl}
Yuan Zhang, Jason Baldridge, and Luheng He. 2019.
\newblock {PAWS: Paraphrase Adversaries from Word Scrambling}.
\newblock In \emph{Proc. of NAACL}.

\bibitem[{Zheng et~al.(2021)Zheng, Guha, Anderson, Henderson, and
  Ho}]{Zheng2021}
Lucia Zheng, Neel Guha, Brandon~R. Anderson, Peter Henderson, and Daniel~E. Ho.
  2021.
\newblock When does pretraining help? assessing self-supervised learning for
  law and the casehold dataset.
\newblock In \emph{International Conference on Artificial Intelligence and
  Law}.

\bibitem[{Zhou et~al.(2022)Zhou, Nie, and Bansal}]{xzhou2022distnli}
Xiang Zhou, Yixin Nie, and Mohit Bansal. 2022.
\newblock Distributed nli: Learning to predict human opinion distributions for
  language reasoning.
\newblock In \emph{Findings of the Association for Computational Linguistics:
  ACL 2022}. Association for Computational Linguistics.

\end{thebibliography}
\bibliographystyle{acl_natbib}
\onecolumn
\appendix
\section{Currently annotated preprocessings \label{sec:appendix}}
\begin{longtable}{lll}
\toprule
{} & preprocessing & task type \\
\midrule
\endfirsthead

\toprule
{} & preprocessing & task type \\
\midrule
\endhead
\midrule
\multicolumn{3}{r}{{Continued on next page}} \\
\midrule
\endfoot

\bottomrule
\endlastfoot
0 & glue/mnli \citep{N18-1101} & Classification \\
1 & glue/qnli \citep{N18-1101} & Classification \\
2 & glue/rte \citep{N18-1101} & Classification \\
3 & glue/wnli \citep{N18-1101} & Classification \\
4 & glue/mrpc \citep{N18-1101} & Classification \\
5 & glue/qqp \citep{N18-1101} & Classification \\
6 & glue/stsb \citep{N18-1101} & Classification \\
7 & super\_glue/boolq \citep{clark2019boolq} & Classification \\
8 & super\_glue/cb \citep{demarneff_simons_tonhauser_2019} & Classification \\
9 & super\_glue/multirc \citep{MultiRC2018} & Classification \\
10 & super\_glue/wic \citep{DBLP:journals/corr/abs-1808-09121} & Classification \\
11 & super\_glue/axg \citep{rudinger-EtAl:2018:N18} & Classification \\
12 & anli/a1 \citep{nie2019adversarial} & Classification \\
13 & anli/a2 \citep{nie2019adversarial} & Classification \\
14 & anli/a3 \citep{nie2019adversarial} & Classification \\
15 & babi\_nli/lists-sets \citep{weston2015towards} & Classification \\
16 & babi\_nli/basic-deduction \citep{weston2015towards} & Classification \\
17 & babi\_nli/positional-reasoning \citep{weston2015towards} & Classification \\
18 & babi\_nli/basic-coreference \citep{weston2015towards} & Classification \\
19 & babi\_nli/three-supporting-facts \citep{weston2015towards} & Classification \\
20 & babi\_nli/path-finding \citep{weston2015towards} & Classification \\
21 & babi\_nli/three-arg-relations \citep{weston2015towards} & Classification \\
22 & babi\_nli/yes-no-questions \citep{weston2015towards} & Classification \\
23 & babi\_nli/time-reasoning \citep{weston2015towards} & Classification \\
24 & babi\_nli/indefinite-knowledge \citep{weston2015towards} & Classification \\
25 & babi\_nli/counting \citep{weston2015towards} & Classification \\
26 & babi\_nli/size-reasoning \citep{weston2015towards} & Classification \\
27 & babi\_nli/compound-coreference \citep{weston2015towards} & Classification \\
28 & babi\_nli/basic-induction \citep{weston2015towards} & Classification \\
29 & babi\_nli/single-supporting-fact \citep{weston2015towards} & Classification \\
30 & babi\_nli/simple-negation \citep{weston2015towards} & Classification \\
31 & babi\_nli/two-arg-relations \citep{weston2015towards} & Classification \\
32 & babi\_nli/two-supporting-facts \citep{weston2015towards} & Classification \\
33 & babi\_nli/conjunction \citep{weston2015towards} & Classification \\
34 & sick/label \citep{marelli-etal-2014-sick} & Classification \\
35 & sick/relatedness \citep{marelli-etal-2014-sick} & Classification \\
36 & sick/entailment\_AB \citep{marelli-etal-2014-sick} & Classification \\
37 & snli \citep{snli:emnlp2015} & Classification \\
38 & scitail/snli\_format \citep{scitail} & Classification \\
39 & hans \citep{DBLP:journals/corr/abs-1902-01007} & Classification \\
40 & WANLI \citep{liu-etal-2022-wanli} & Classification \\
41 & recast/recast\_verbcorner \citep{poliak-etal-2018-collecting} & Classification \\
42 & recast/recast\_megaveridicality \citep{poliak-etal-2018-collecting} & Classification \\
43 & recast/recast\_sentiment \citep{poliak-etal-2018-collecting} & Classification \\
44 & recast/recast\_ner \citep{poliak-etal-2018-collecting} & Classification \\
45 & recast/recast\_kg\_relations \citep{poliak-etal-2018-collecting} & Classification \\
46 & recast/recast\_factuality \citep{poliak-etal-2018-collecting} & Classification \\
47 & recast/recast\_puns \citep{poliak-etal-2018-collecting} & Classification \\
48 & recast/recast\_verbnet \citep{poliak-etal-2018-collecting} & Classification \\
49 & probability\_words\_nli/reasoning\_1hop \citep{sileo2022probing} & Classification \\
50 & probability\_words\_nli/usnli \citep{sileo2022probing} & Classification \\
51 & probability\_words\_nli/reasoning\_2hop \citep{sileo2022probing} & Classification \\
52 & nan-nli/joey234--nan-nli & Classification \\
53 & nli\_fever & Classification \\
54 & breaking\_nli & Classification \\
55 & conj\_nli & Classification \\
56 & fracas & Classification \\
57 & dialogue\_nli & Classification \\
58 & mpe & Classification \\
59 & dnc & Classification \\
60 & recast\_white/fnplus & Classification \\
61 & recast\_white/sprl & Classification \\
62 & recast\_white/dpr & Classification \\
63 & joci & Classification \\
64 & robust\_nli/IS\_CS & Classification \\
65 & robust\_nli/LI\_LI & Classification \\
66 & robust\_nli/ST\_WO & Classification \\
67 & robust\_nli/PI\_SP & Classification \\
68 & robust\_nli/PI\_CD & Classification \\
69 & robust\_nli/ST\_SE & Classification \\
70 & robust\_nli/ST\_NE & Classification \\
71 & robust\_nli/ST\_LM & Classification \\
72 & robust\_nli\_is\_sd & Classification \\
73 & robust\_nli\_li\_ts & Classification \\
74 & gen\_debiased\_nli/snli\_seq\_z & Classification \\
75 & gen\_debiased\_nli/snli\_z\_aug & Classification \\
76 & gen\_debiased\_nli/snli\_par\_z & Classification \\
77 & gen\_debiased\_nli/mnli\_par\_z & Classification \\
78 & gen\_debiased\_nli/mnli\_z\_aug & Classification \\
79 & gen\_debiased\_nli/mnli\_seq\_z & Classification \\
80 & add\_one\_rte & Classification \\
81 & imppres/presupposition\_all\_n\_presupposition \citep{jeretic-etal-2020-natural} & Classification \\
82 & imppres/presupposition\_possessed\_definites\_existence \citep{jeretic-etal-2020-natural} & Classification \\
83 & imppres/presupposition\_cleft\_uniqueness\citep{jeretic-etal-2020-natural} & Classification \\
84 & imppres/presupposition\_question\_presupposition\citep{jeretic-etal-2020-natural} & Classification \\
85 & imppres/presupposition\_possessed\_definites\_uniqueness\citep{jeretic-etal-2020-natural} & Classification \\
86 & imppres/presupposition\_only\_presupposition\citep{jeretic-etal-2020-natural} & Classification \\
87 & imppres/presupposition\_both\_presupposition\citep{jeretic-etal-2020-natural} & Classification \\
88 & imppres/presupposition\_change\_of\_state\citep{jeretic-etal-2020-natural} & Classification \\
89 & imppres/presupposition\_cleft\_existence\citep{jeretic-etal-2020-natural} & Classification \\
90 & imppres/implicature\_quantifiers/prag \citep{jeretic-etal-2020-natural} & Classification \\
91 & imppres/implicature\_numerals\_2\_3/prag \citep{jeretic-etal-2020-natural} & Classification \\
92 & imppres/implicature\_numerals\_10\_100/prag \citep{jeretic-etal-2020-natural} & Classification \\
93 & imppres/implicature\_modals/prag \citep{jeretic-etal-2020-natural} & Classification \\
94 & imppres/implicature\_connectives/prag \citep{jeretic-etal-2020-natural} & Classification \\
95 & imppres/implicature\_gradable\_verb/prag \citep{jeretic-etal-2020-natural} & Classification \\
96 & imppres/implicature\_gradable\_adjective/prag \citep{jeretic-etal-2020-natural} & Classification \\
97 & imppres/implicature\_quantifiers/log \citep{jeretic-etal-2020-natural} & Classification \\
98 & imppres/implicature\_numerals\_2\_3/log \citep{jeretic-etal-2020-natural} & Classification \\
99 & imppres/implicature\_numerals\_10\_100/log \citep{jeretic-etal-2020-natural} & Classification \\
100 & imppres/implicature\_gradable\_adjective/log \citep{jeretic-etal-2020-natural} & Classification \\
101 & imppres/implicature\_connectives/log \citep{jeretic-etal-2020-natural} & Classification \\
102 & imppres/implicature\_modals/log \citep{jeretic-etal-2020-natural} & Classification \\
103 & imppres/implicature\_gradable\_verb/log \citep{jeretic-etal-2020-natural} & Classification \\
104 & glue\_diagnostics/diagnostics & Classification \\
105 & hlgd \citep{Laban2021NewsHG} & Classification \\
106 & paws/labeled\_final \citep{paws2019naacl} & Classification \\
107 & paws/labeled\_swap \citep{paws2019naacl} & Classification \\
108 & quora & Classification \\
109 & medical\_questions\_pairs \citep{mccreery2020effective} & Classification \\
110 & glue/cola \citep{N18-1101} & Classification \\
111 & glue/sst2 \citep{N18-1101} & Classification \\
112 & utilitarianism \citep{hendrycks2020ethics} & Classification \\
113 & amazon\_counterfactual/en \citep{oneill2021i} & Classification \\
114 & insincere-questions & Classification \\
115 & toxic\_conversations & Classification \\
116 & TuringBench \citep{huggingface:dataset} & Classification \\
117 & trec \citep{li-roth-2002-learning} & Classification \\
118 & vitaminc/tals--vitaminc \citep{schuster-etal-2021-get} & Classification \\
119 & hope\_edi/english \citep{chakravarthi-2020-hopeedi} & Classification \\
120 & rumoureval\_2019/RumourEval2019 \citep{gorrell-etal-2019-semeval} & Classification \\
121 & ethos/binary \citep{mollas2020ethos} & Classification \\
122 & ethos/multilabel \citep{mollas2020ethos} & Classification \\
123 & tweet\_eval/emotion \citep{barbieri2020tweeteval} & Classification \\
124 & tweet\_eval/irony \citep{barbieri2020tweeteval} & Classification \\
125 & tweet\_eval/offensive \citep{barbieri2020tweeteval} & Classification \\
126 & tweet\_eval/sentiment \citep{barbieri2020tweeteval} & Classification \\
127 & tweet\_eval/stance\_abortion \citep{barbieri2020tweeteval} & Classification \\
128 & tweet\_eval/stance\_atheism \citep{barbieri2020tweeteval} & Classification \\
129 & tweet\_eval/stance\_climate \citep{barbieri2020tweeteval} & Classification \\
130 & tweet\_eval/stance\_feminist \citep{barbieri2020tweeteval} & Classification \\
131 & tweet\_eval/stance\_hillary \citep{barbieri2020tweeteval} & Classification \\
132 & tweet\_eval/emoji \citep{barbieri2020tweeteval} & Classification \\
133 & tweet\_eval/hate \citep{barbieri2020tweeteval} & Classification \\
134 & discovery/discovery \citep{sileo-etal-2019-mining} & Classification \\
135 & pragmeval/squinky-informativeness \citep{DBLP:journals/corr/Lahiri15} & Classification \\
136 & pragmeval/squinky-implicature \citep{DBLP:journals/corr/Lahiri15} & Classification \\
137 & pragmeval/verifiability \citep{park2014identifying} & Classification \\
138 & pragmeval/squinky-formality \citep{DBLP:journals/corr/Lahiri15} & Classification \\
139 & pragmeval/emobank-valence \citep{buechel-hahn-2017-emobank} & Classification \\
140 & pragmeval/emobank-dominance \citep{buechel-hahn-2017-emobank} & Classification \\
141 & pragmeval/emobank-arousal \citep{buechel-hahn-2017-emobank} & Classification \\
142 & pragmeval/switchboard \citep{Godfrey:1992:STS:1895550.1895693} & Classification \\
143 & pragmeval/mrda \citep{shriberg2004icsi} & Classification \\
144 & pragmeval/sarcasm \citep{OrabySarc} & Classification \\
145 & pragmeval/persuasiveness-premisetype \citep{Persuasion2018Ng} & Classification \\
146 & pragmeval/persuasiveness-eloquence \citep{Persuasion2018Ng} & Classification \\
147 & pragmeval/persuasiveness-claimtype \citep{Persuasion2018Ng} & Classification \\
148 & pragmeval/persuasiveness-specificity \citep{Persuasion2018Ng} & Classification \\
149 & pragmeval/gum \citep{Zeldes2017} & Classification \\
150 & pragmeval/emergent \citep{Ferreira2016EmergentAN} & Classification \\
151 & pragmeval/persuasiveness-strength \citep{Persuasion2018Ng} & Classification \\
152 & pragmeval/stac \citep{asher-etal-2016-discourse} & Classification \\
153 & pragmeval/pdtb \citep{prasad-etal-2008-penn} & Classification \\
154 & pragmeval/persuasiveness-relevance \citep{Persuasion2018Ng} & Classification \\
155 & silicone/meld\_s \citep{chen2018emotionlines} & Classification \\
156 & silicone/sem \citep{mckeown2011semaine} & Classification \\
157 & silicone/oasis \citep{leech2003generic} & Classification \\
158 & silicone/meld\_e \citep{chen2018emotionlines} & Classification \\
159 & silicone/maptask \citep{thompson1993hcrc} & Classification \\
160 & silicone/iemocap \citep{busso2008iemocap} & Classification \\
161 & silicone/dyda\_e \citep{li2017dailydialog} & Classification \\
162 & silicone/dyda\_da \citep{li2017dailydialog} & Classification \\
163 & lex\_glue/eurlex \citep{chalkidis-etal-2021-multieurlex} & Classification \\
164 & lex\_glue/scotus \citep{spaeth2020} & Classification \\
165 & lex\_glue/ledgar \citep{tuggener-etal-2020-ledgar} & Classification \\
166 & lex\_glue/unfair\_tos \citep{lippi-etal-2019-claudette} & Classification \\
167 & language-identification & Classification \\
168 & imdb \citep{maas-EtAl:2011:ACL-HLT2011} & Classification \\
169 & rotten\_tomatoes \citep{Pang+Lee:05a} & Classification \\
170 & ag\_news \citep{Zhang2015CharacterlevelCN} & Classification \\
171 & yelp\_review\_full/yelp\_review\_full \citep{zhang2015character} & Classification \\
172 & financial\_phrasebank/sentences\_allagree \citep{Malo2014GoodDO} & Classification \\
173 & poem\_sentiment \citep{sheng2020investigating} & Classification \\
174 & dbpedia\_14/dbpedia\_14 \citep{lehmann2015dbpedia} & Classification \\
175 & amazon\_polarity/amazon\_polarity \citep{mcauley2013hidden} & Classification \\
176 & app\_reviews \citep{ZurichOpenRepositoryandArchive:dataset} & Classification \\
177 & hate\_speech18 \citep{gibert2018hate} & Classification \\
178 & sms\_spam \citep{Almeida2011SpamFiltering} & Classification \\
179 & humicroedit/subtask-1 \citep{hossain2019president} & Classification \\
180 & humicroedit/subtask-2 \citep{hossain2019president} & Classification \\
181 & snips\_built\_in\_intents \citep{DBLP:journals/corr/abs-1805-10190} & Classification \\
182 & banking77 \citep{Casanueva2020} & Classification \\
183 & hate\_speech\_offensive \citep{hateoffensive} & Classification \\
184 & yahoo\_answers\_topics & Classification \\
185 & stackoverflow-questions & Classification \\
186 & hyperpartisan\_news & Classification \\
187 & sciie & Classification \\
188 & citation\_intent & Classification \\
189 & go\_emotions/simplified \citep{demszky2020goemotions} & Classification \\
190 & scicite \citep{Cohan2019Structural} & Classification \\
191 & liar \citep{wang-2017-liar} & Classification \\
192 & lexical\_relation\_classification/K\&H+N \citep{wang-etal-2019-spherere} & Classification \\
193 & lexical\_relation\_classification/CogALexV \citep{wang-etal-2019-spherere} & Classification \\
194 & lexical\_relation\_classification/BLESS \citep{wang-etal-2019-spherere} & Classification \\
195 & lexical\_relation\_classification/EVALution \citep{wang-etal-2019-spherere} & Classification \\
196 & lexical\_relation\_classification/ROOT09 \citep{wang-etal-2019-spherere} & Classification \\
197 & linguisticprobing/subj\_number \citep{conneau-etal-2018-cram} & Classification \\
198 & linguisticprobing/bigram\_shift \citep{conneau-etal-2018-cram} & Classification \\
199 & linguisticprobing/top\_constituents \citep{conneau-etal-2018-cram} & Classification \\
200 & linguisticprobing/odd\_man\_out \citep{conneau-etal-2018-cram} & Classification \\
201 & linguisticprobing/past\_present \citep{conneau-etal-2018-cram} & Classification \\
202 & linguisticprobing/coordination\_inversion \citep{conneau-etal-2018-cram} & Classification \\
203 & linguisticprobing/tree\_depth \citep{conneau-etal-2018-cram} & Classification \\
204 & linguisticprobing/obj\_number \citep{conneau-etal-2018-cram} & Classification \\
205 & linguisticprobing/sentence\_length \citep{conneau-etal-2018-cram} & Classification \\
206 & crowdflower/sentiment\_nuclear\_power \citep{van2012designing} & Classification \\
207 & crowdflower/tweet\_global\_warming \citep{van2012designing} & Classification \\
208 & crowdflower/corporate-messaging \citep{van2012designing} & Classification \\
209 & crowdflower/economic-news \citep{van2012designing} & Classification \\
210 & crowdflower/airline-sentiment \citep{van2012designing} & Classification \\
211 & crowdflower/political-media-bias \citep{van2012designing} & Classification \\
212 & crowdflower/text\_emotion \citep{van2012designing} & Classification \\
213 & crowdflower/political-media-audience \citep{van2012designing} & Classification \\
214 & crowdflower/political-media-message \citep{van2012designing} & Classification \\
215 & ethics/commonsense \citep{hendrycks2020ethics} & Classification \\
216 & ethics/deontology \citep{hendrycks2020ethics} & Classification \\
217 & ethics/justice \citep{hendrycks2020ethics} & Classification \\
218 & ethics/virtue \citep{hendrycks2020ethics} & Classification \\
219 & emo/emo2019 \citep{chatterjee-etal-2019-semeval} & Classification \\
220 & google\_wellformed\_query \citep{faruqui2018identifying} & Classification \\
221 & tweets\_hate\_speech\_detection \citep{ZRoshanSharma:dataset} & Classification \\
222 & has\_part \citep{bhakthavatsalam2020dogs} & Classification \\
223 & blog\_authorship\_corpus/gender \citep{schler2006effects} & Classification \\
224 & blog\_authorship\_corpus/age \citep{schler2006effects} & Classification \\
225 & blog\_authorship\_corpus/horoscope \citep{schler2006effects} & Classification \\
226 & blog\_authorship\_corpus/job \citep{schler2006effects} & Classification \\
227 & open\_question\_type \citep{cao-wang-2021-controllable} & Classification \\
228 & health\_fact \citep{kotonya-toni-2020-explainable} & Classification \\
229 & mc\_taco \citep{ZKNR19} & Classification \\
230 & ade\_corpus\_v2/Ade\_corpus\_v2\_classification \citep{GURULINGAPPA2012885} & Classification \\
231 & circa \citep{louis_emnlp2020} & Classification \\
232 & EffectiveFeedbackStudentWriting & Classification \\
233 & promptSentiment \citep{mcauley2013hidden} & Classification \\
234 & promptNLI \citep{nie2019adversarial} & Classification \\
235 & promptSpoke & Classification \\
236 & promptProficiency & Classification \\
237 & promptGrammar \citep{warstadt2018neural} & Classification \\
238 & promptCoherence & Classification \\
239 & phrase\_similarity \citep{pham2022PiC} & Classification \\
240 & scientific-exaggeration-detection \citep{wright2021exaggeration} & Classification \\
241 & quarel & Classification \\
242 & fever-evidence-related/mwong--fever-related & Classification \\
243 & numer\_sense \citep{lin2020numersense} & Classification \\
244 & dynasent/dynabench.dynasent.r1.all/r1 \citep{potts-etal-2020-dynasent} & Classification \\
245 & dynasent/dynabench.dynasent.r2.all/r2 \citep{potts-etal-2020-dynasent} & Classification \\
246 & Sarcasm\_News\_Headline & Classification \\
247 & sem\_eval\_2010\_task\_8 \citep{hendrickx-etal-2010-semeval} & Classification \\
248 & auditor\_review/demo-org--auditor\_review & Classification \\
249 & Dynasent\_Disagreement & Classification \\
250 & Politeness\_Disagreement & Classification \\
251 & SBIC\_Disagreement & Classification \\
252 & SChem\_Disagreement & Classification \\
253 & Dilemmas\_Disagreement & Classification \\
254 & wiki\_qa \citep{YangYihMeek:EMNLP2015:WikiQA} & Classification \\
255 & cycic\_classification \citep{Kejriwal2020DoFC} & Classification \\
256 & sts-companion \citep{cer-etal-2017-semeval} & Classification \\
257 & commonsense\_qa\_2.0 & Classification \\
258 & lingnli \citep{parrish-etal-2021-putting-linguist} & Classification \\
259 & monotonicity-entailment \citep{yanaka-etal-2019-neural} & Classification \\
260 & scinli \citep{sadat-caragea-2022-scinli} & Classification \\
261 & naturallogic \citep{feng2020exploring} & Classification \\
262 & dynahate \citep{vidgen2021learning} & Classification \\
263 & syntactic-augmentation-nli \citep{min-etal-2020-syntactic} & Classification \\
264 & autotnli & Classification \\
265 & CONDAQA \citep{ravichander-et-al-2022-condaqa} & Classification \\
266 & scruples & Classification \\
267 & attempto-nli & Classification \\
268 & defeasible-nli/atomic & Classification \\
269 & defeasible-nli/snli & Classification \\
270 & help-nli \citep{yanaka-EtAl:2019:starsem} & Classification \\
271 & nli-veridicality-transitivity \citep{yanaka-etal-2021-exploring} & Classification \\
272 & natural-language-satisfiability \citep{https://doi.org/10.48550/arxiv.2211.05417} & Classification \\
273 & lonli \citep{Tarunesh2021TrustingRO} & Classification \\
274 & dadc-limit-nli \citep{Wallace2022Dynamic} & Classification \\
275 & FLUTE & Classification \\
276 & strategy-qa & Classification \\
277 & folio \citep{han2022folio} & Classification \\
278 & tomi-nli & Classification \\
279 & avicenna \citep{aghahadi2022avicenna} & Classification \\
280 & CREAK & Classification \\
281 & puzzte \citep{szomiu2021puzzle} & Classification \\
282 & spartqa-yn \citep{mirzaee-etal-2021-spartqa} & Classification \\
283 & temporal-nli \citep{thukral-etal-2021-probing} & Classification \\
284 & clcd-english & Classification \\
285 & twentyquestions & Classification \\
286 & counterfactually-augmented-imdb \citep{kaushik2020learning} & Classification \\
287 & counterfactually-augmented-snli \citep{kaushik2020learning} & Classification \\
288 & cnli \citep{huang2020cnligeneralization} & Classification \\
289 & boolq-natural-perturbations \citep{khashabi2020naturalperturbations} & Classification \\
290 & acceptability-prediction \citep{lau-etal-2015-unsupervised} & Classification \\
291 & equate \citep{ravichander2019equate} & Classification \\
292 & implicit-hate-stg1 \citep{elsherief-etal-2021-latent} & Classification \\
293 & chaos-mnli-ambiguity \citep{xzhou2022distnli} & Classification \\
294 & headline\_cause/en\_simple \citep{gusev2021headlinecause} & Classification \\
295 & logiqa-2.0-nli & Classification \\
296 & oasst1\_dense\_flat/quality & Classification \\
297 & oasst1\_dense\_flat/toxicity & Classification \\
298 & oasst1\_dense\_flat/helpfulness & Classification \\
299 & PARARULE-Plus \citep{bao2022multi} & Classification \\
300 & mindgames \citep{sileo2023mindgames} & Classification \\
301 & ambient \citep{liu-etal-2023-afraid} & Classification \\
302 & civil\_comments/toxicity \citep{DBLP:journals/corr/abs-1903-04561} & Classification \\
303 & civil\_comments/severe\_toxicity \citep{DBLP:journals/corr/abs-1903-04561} & Classification \\
304 & civil\_comments/obscene \citep{DBLP:journals/corr/abs-1903-04561} & Classification \\
305 & civil\_comments/threat \citep{DBLP:journals/corr/abs-1903-04561} & Classification \\
306 & civil\_comments/insult \citep{DBLP:journals/corr/abs-1903-04561} & Classification \\
307 & civil\_comments/identity\_attack \citep{DBLP:journals/corr/abs-1903-04561} & Classification \\
308 & civil\_comments/sexual\_explicit \citep{DBLP:journals/corr/abs-1903-04561} & Classification \\
309 & I2D2 & Classification \\
310 & hh-rlhf & MultipleChoice \\
311 & model-written-evals \citep{perez2022discovering} & MultipleChoice \\
312 & truthful\_qa/multiple\_choice \citep{lin2021truthfulqa} & MultipleChoice \\
313 & fig-qa & MultipleChoice \\
314 & bigbench/strange\_stories \citep{srivastava2022beyond} & MultipleChoice \\
315 & bigbench/arithmetic \citep{srivastava2022beyond} & MultipleChoice \\
316 & bigbench/formal\_fallacies\_syllogisms\_negation \citep{srivastava2022beyond} & MultipleChoice \\
317 & bigbench/implicatures \citep{srivastava2022beyond} & MultipleChoice \\
318 & bigbench/salient\_translation\_error\_detection \citep{srivastava2022beyond} & MultipleChoice \\
319 & bigbench/causal\_judgment \citep{srivastava2022beyond} & MultipleChoice \\
320 & bigbench/discourse\_marker\_prediction \citep{srivastava2022beyond} & MultipleChoice \\
321 & bigbench/timedial \citep{srivastava2022beyond} & MultipleChoice \\
322 & bigbench/general\_knowledge \citep{srivastava2022beyond} & MultipleChoice \\
323 & bigbench/evaluating\_information\_essentiality \citep{srivastava2022beyond} & MultipleChoice \\
324 & bigbench/cause\_and\_effect \citep{srivastava2022beyond} & MultipleChoice \\
325 & bigbench/hyperbaton \citep{srivastava2022beyond} & MultipleChoice \\
326 & bigbench/hindu\_knowledge \citep{srivastava2022beyond} & MultipleChoice \\
327 & bigbench/crass\_ai \citep{srivastava2022beyond} & MultipleChoice \\
328 & bigbench/movie\_recommendation \citep{srivastava2022beyond} & MultipleChoice \\
329 & bigbench/cifar10\_classification \citep{srivastava2022beyond} & MultipleChoice \\
330 & bigbench/logic\_grid\_puzzle \citep{srivastava2022beyond} & MultipleChoice \\
331 & bigbench/sentence\_ambiguity \citep{srivastava2022beyond} & MultipleChoice \\
332 & bigbench/fact\_checker \citep{srivastava2022beyond} & MultipleChoice \\
333 & bigbench/strategyqa \citep{srivastava2022beyond} & MultipleChoice \\
334 & bigbench/elementary\_math\_qa \citep{srivastava2022beyond} & MultipleChoice \\
335 & bigbench/temporal\_sequences \citep{srivastava2022beyond} & MultipleChoice \\
336 & bigbench/penguins\_in\_a\_table \citep{srivastava2022beyond} & MultipleChoice \\
337 & bigbench/goal\_step\_wikihow \citep{srivastava2022beyond} & MultipleChoice \\
338 & bigbench/dark\_humor\_detection \citep{srivastava2022beyond} & MultipleChoice \\
339 & bigbench/logical\_fallacy\_detection \citep{srivastava2022beyond} & MultipleChoice \\
340 & bigbench/irony\_identification \citep{srivastava2022beyond} & MultipleChoice \\
341 & bigbench/emojis\_emotion\_prediction \citep{srivastava2022beyond} & MultipleChoice \\
342 & bigbench/sports\_understanding \citep{srivastava2022beyond} & MultipleChoice \\
343 & bigbench/contextual\_parametric\_knowledge\_conflicts \citep{srivastava2022beyond} & MultipleChoice \\
344 & bigbench/intent\_recognition \citep{srivastava2022beyond} & MultipleChoice \\
345 & bigbench/crash\_blossom \citep{srivastava2022beyond} & MultipleChoice \\
346 & bigbench/real\_or\_fake\_text \citep{srivastava2022beyond} & MultipleChoice \\
347 & bigbench/ruin\_names \citep{srivastava2022beyond} & MultipleChoice \\
348 & bigbench/logical\_deduction \citep{srivastava2022beyond} & MultipleChoice \\
349 & bigbench/identify\_math\_theorems \citep{srivastava2022beyond} & MultipleChoice \\
350 & bigbench/vitaminc\_fact\_verification \citep{srivastava2022beyond} & MultipleChoice \\
351 & bigbench/hhh\_alignment \citep{srivastava2022beyond} & MultipleChoice \\
352 & bigbench/simple\_ethical\_questions \citep{srivastava2022beyond} & MultipleChoice \\
353 & bigbench/checkmate\_in\_one \citep{srivastava2022beyond} & MultipleChoice \\
354 & bigbench/similarities\_abstraction \citep{srivastava2022beyond} & MultipleChoice \\
355 & bigbench/novel\_concepts \citep{srivastava2022beyond} & MultipleChoice \\
356 & bigbench/snarks \citep{srivastava2022beyond} & MultipleChoice \\
357 & bigbench/abstract\_narrative\_understanding \citep{srivastava2022beyond} & MultipleChoice \\
358 & bigbench/social\_iqa \citep{srivastava2022beyond} & MultipleChoice \\
359 & bigbench/phrase\_relatedness \citep{srivastava2022beyond} & MultipleChoice \\
360 & bigbench/physics \citep{srivastava2022beyond} & MultipleChoice \\
361 & bigbench/gre\_reading\_comprehension \citep{srivastava2022beyond} & MultipleChoice \\
362 & bigbench/logical\_sequence \citep{srivastava2022beyond} & MultipleChoice \\
363 & bigbench/winowhy \citep{srivastava2022beyond} & MultipleChoice \\
364 & bigbench/movie\_dialog\_same\_or\_different \citep{srivastava2022beyond} & MultipleChoice \\
365 & bigbench/riddle\_sense \citep{srivastava2022beyond} & MultipleChoice \\
366 & bigbench/metaphor\_understanding \citep{srivastava2022beyond} & MultipleChoice \\
367 & bigbench/moral\_permissibility \citep{srivastava2022beyond} & MultipleChoice \\
368 & bigbench/nonsense\_words\_grammar \citep{srivastava2022beyond} & MultipleChoice \\
369 & bigbench/bbq\_lite\_json \citep{srivastava2022beyond} & MultipleChoice \\
370 & bigbench/physical\_intuition \citep{srivastava2022beyond} & MultipleChoice \\
371 & bigbench/navigate \citep{srivastava2022beyond} & MultipleChoice \\
372 & bigbench/reasoning\_about\_colored\_objects \citep{srivastava2022beyond} & MultipleChoice \\
373 & bigbench/metaphor\_boolean \citep{srivastava2022beyond} & MultipleChoice \\
374 & bigbench/analytic\_entailment \citep{srivastava2022beyond} & MultipleChoice \\
375 & bigbench/mnist\_ascii \citep{srivastava2022beyond} & MultipleChoice \\
376 & bigbench/misconceptions \citep{srivastava2022beyond} & MultipleChoice \\
377 & bigbench/authorship\_verification \citep{srivastava2022beyond} & MultipleChoice \\
378 & bigbench/social\_support \citep{srivastava2022beyond} & MultipleChoice \\
379 & bigbench/tracking\_shuffled\_objects \citep{srivastava2022beyond} & MultipleChoice \\
380 & bigbench/analogical\_similarity \citep{srivastava2022beyond} & MultipleChoice \\
381 & bigbench/figure\_of\_speech\_detection \citep{srivastava2022beyond} & MultipleChoice \\
382 & bigbench/understanding\_fables \citep{srivastava2022beyond} & MultipleChoice \\
383 & bigbench/question\_selection \citep{srivastava2022beyond} & MultipleChoice \\
384 & bigbench/undo\_permutation \citep{srivastava2022beyond} & MultipleChoice \\
385 & bigbench/conceptual\_combinations \citep{srivastava2022beyond} & MultipleChoice \\
386 & bigbench/unit\_interpretation \citep{srivastava2022beyond} & MultipleChoice \\
387 & bigbench/logical\_args \citep{srivastava2022beyond} & MultipleChoice \\
388 & bigbench/geometric\_shapes \citep{srivastava2022beyond} & MultipleChoice \\
389 & bigbench/code\_line\_description \citep{srivastava2022beyond} & MultipleChoice \\
390 & bigbench/fantasy\_reasoning \citep{srivastava2022beyond} & MultipleChoice \\
391 & bigbench/identify\_odd\_metaphor \citep{srivastava2022beyond} & MultipleChoice \\
392 & bigbench/empirical\_judgments \citep{srivastava2022beyond} & MultipleChoice \\
393 & bigbench/color \citep{srivastava2022beyond} & MultipleChoice \\
394 & bigbench/symbol\_interpretation \citep{srivastava2022beyond} & MultipleChoice \\
395 & bigbench/suicide\_risk \citep{srivastava2022beyond} & MultipleChoice \\
396 & bigbench/date\_understanding \citep{srivastava2022beyond} & MultipleChoice \\
397 & bigbench/cs\_algorithms \citep{srivastava2022beyond} & MultipleChoice \\
398 & bigbench/play\_dialog\_same\_or\_different \citep{srivastava2022beyond} & MultipleChoice \\
399 & bigbench/international\_phonetic\_alphabet\_nli \citep{srivastava2022beyond} & MultipleChoice \\
400 & bigbench/emoji\_movie \citep{srivastava2022beyond} & MultipleChoice \\
401 & bigbench/mathematical\_induction \citep{srivastava2022beyond} & MultipleChoice \\
402 & bigbench/implicit\_relations \citep{srivastava2022beyond} & MultipleChoice \\
403 & bigbench/anachronisms \citep{srivastava2022beyond} & MultipleChoice \\
404 & bigbench/odd\_one\_out \citep{srivastava2022beyond} & MultipleChoice \\
405 & bigbench/human\_organs\_senses \citep{srivastava2022beyond} & MultipleChoice \\
406 & bigbench/english\_proverbs \citep{srivastava2022beyond} & MultipleChoice \\
407 & bigbench/key\_value\_maps \citep{srivastava2022beyond} & MultipleChoice \\
408 & bigbench/dyck\_languages \citep{srivastava2022beyond} & MultipleChoice \\
409 & bigbench/known\_unknowns \citep{srivastava2022beyond} & MultipleChoice \\
410 & bigbench/disambiguation\_qa \citep{srivastava2022beyond} & MultipleChoice \\
411 & bigbench/entailed\_polarity \citep{srivastava2022beyond} & MultipleChoice \\
412 & bigbench/epistemic\_reasoning \citep{srivastava2022beyond} & MultipleChoice \\
413 & bigbench/presuppositions\_as\_nli \citep{srivastava2022beyond} & MultipleChoice \\
414 & blimp/sentential\_negation\_npi\_scope \citep{warstadt2019blimp} & MultipleChoice \\
415 & blimp/left\_branch\_island\_echo\_question \citep{warstadt2019blimp} & MultipleChoice \\
416 & blimp/inchoative \citep{warstadt2019blimp} & MultipleChoice \\
417 & blimp/principle\_A\_reconstruction \citep{warstadt2019blimp} & MultipleChoice \\
418 & blimp/complex\_NP\_island \citep{warstadt2019blimp} & MultipleChoice \\
419 & blimp/npi\_present\_2 \citep{warstadt2019blimp} & MultipleChoice \\
420 & blimp/existential\_there\_quantifiers\_2 \citep{warstadt2019blimp} & MultipleChoice \\
421 & blimp/wh\_vs\_that\_with\_gap \citep{warstadt2019blimp} & MultipleChoice \\
422 & blimp/superlative\_quantifiers\_1 \citep{warstadt2019blimp} & MultipleChoice \\
423 & blimp/coordinate\_structure\_constraint\_complex\_left\_branch \citep{warstadt2019blimp} & MultipleChoice \\
424 & blimp/matrix\_question\_npi\_licensor\_present \citep{warstadt2019blimp} & MultipleChoice \\
425 & blimp/principle\_A\_c\_command \citep{warstadt2019blimp} & MultipleChoice \\
426 & blimp/drop\_argument \citep{warstadt2019blimp} & MultipleChoice \\
427 & blimp/tough\_vs\_raising\_1 \citep{warstadt2019blimp} & MultipleChoice \\
428 & blimp/npi\_present\_1 \citep{warstadt2019blimp} & MultipleChoice \\
429 & blimp/coordinate\_structure\_constraint\_object\_extraction \citep{warstadt2019blimp} & MultipleChoice \\
430 & blimp/animate\_subject\_passive \citep{warstadt2019blimp} & MultipleChoice \\
431 & blimp/wh\_vs\_that\_with\_gap\_long\_distance \citep{warstadt2019blimp} & MultipleChoice \\
432 & blimp/wh\_questions\_subject\_gap\_long\_distance \citep{warstadt2019blimp} & MultipleChoice \\
433 & blimp/sentential\_subject\_island \citep{warstadt2019blimp} & MultipleChoice \\
434 & blimp/wh\_questions\_object\_gap \citep{warstadt2019blimp} & MultipleChoice \\
435 & blimp/principle\_A\_domain\_2 \citep{warstadt2019blimp} & MultipleChoice \\
436 & cos\_e/v1.0 \citep{rajani2019explain} & MultipleChoice \\
437 & cosmos\_qa \citep{huang-etal-2019-cosmos} & MultipleChoice \\
438 & dream \citep{sundream2018} & MultipleChoice \\
439 & openbookqa \citep{OpenBookQA2018} & MultipleChoice \\
440 & qasc \citep{allenai:qasc} & MultipleChoice \\
441 & quartz \citep{quartz} & MultipleChoice \\
442 & quail \citep{DBLP:conf/aaai/RogersKDR20} & MultipleChoice \\
443 & head\_qa/en \citep{vilares-gomez-rodriguez-2019-head} & MultipleChoice \\
444 & sciq \citep{SciQ} & MultipleChoice \\
445 & social\_i\_qa & MultipleChoice \\
446 & wiki\_hop/original \citep{welbl2018constructing} & MultipleChoice \\
447 & wiqa \citep{wiqa} & MultipleChoice \\
448 & piqa \citep{Bisk2020} & MultipleChoice \\
449 & hellaswag \citep{zellers2019hellaswag} & MultipleChoice \\
450 & super\_glue/copa \citep{roemmele2011choice} & MultipleChoice \\
451 & balanced-copa \citep{kavumba-etal-2019-choosing} & MultipleChoice \\
452 & e-CARE & MultipleChoice \\
453 & art \citep{anli} & MultipleChoice \\
454 & mmlu/nutrition \citep{hendryckstest2021} & MultipleChoice \\
455 & mmlu/college\_medicine \citep{hendryckstest2021} & MultipleChoice \\
456 & mmlu/philosophy \citep{hendryckstest2021} & MultipleChoice \\
457 & mmlu/global\_facts \citep{hendryckstest2021} & MultipleChoice \\
458 & mmlu/college\_mathematics \citep{hendryckstest2021} & MultipleChoice \\
459 & mmlu/college\_computer\_science \citep{hendryckstest2021} & MultipleChoice \\
460 & mmlu/college\_chemistry \citep{hendryckstest2021} & MultipleChoice \\
461 & mmlu/college\_biology \citep{hendryckstest2021} & MultipleChoice \\
462 & mmlu/clinical\_knowledge \citep{hendryckstest2021} & MultipleChoice \\
463 & mmlu/business\_ethics \citep{hendryckstest2021} & MultipleChoice \\
464 & mmlu/astronomy \citep{hendryckstest2021} & MultipleChoice \\
465 & mmlu/machine\_learning \citep{hendryckstest2021} & MultipleChoice \\
466 & mmlu/moral\_scenarios \citep{hendryckstest2021} & MultipleChoice \\
467 & mmlu/sociology \citep{hendryckstest2021} & MultipleChoice \\
468 & mmlu/us\_foreign\_policy \citep{hendryckstest2021} & MultipleChoice \\
469 & mmlu/virology \citep{hendryckstest2021} & MultipleChoice \\
470 & mmlu/world\_religions \citep{hendryckstest2021} & MultipleChoice \\
471 & mmlu/prehistory \citep{hendryckstest2021} & MultipleChoice \\
472 & mmlu/professional\_accounting \citep{hendryckstest2021} & MultipleChoice \\
473 & mmlu/professional\_law \citep{hendryckstest2021} & MultipleChoice \\
474 & mmlu/professional\_medicine \citep{hendryckstest2021} & MultipleChoice \\
475 & mmlu/professional\_psychology \citep{hendryckstest2021} & MultipleChoice \\
476 & mmlu/electrical\_engineering \citep{hendryckstest2021} & MultipleChoice \\
477 & mmlu/elementary\_mathematics \citep{hendryckstest2021} & MultipleChoice \\
478 & mmlu/anatomy \citep{hendryckstest2021} & MultipleChoice \\
479 & mmlu/abstract\_algebra \citep{hendryckstest2021} & MultipleChoice \\
480 & mmlu/medical\_genetics \citep{hendryckstest2021} & MultipleChoice \\
481 & mmlu/miscellaneous \citep{hendryckstest2021} & MultipleChoice \\
482 & mmlu/logical\_fallacies \citep{hendryckstest2021} & MultipleChoice \\
483 & mmlu/jurisprudence \citep{hendryckstest2021} & MultipleChoice \\
484 & mmlu/computer\_security \citep{hendryckstest2021} & MultipleChoice \\
485 & mmlu/international\_law \citep{hendryckstest2021} & MultipleChoice \\
486 & mmlu/human\_sexuality \citep{hendryckstest2021} & MultipleChoice \\
487 & mmlu/human\_aging \citep{hendryckstest2021} & MultipleChoice \\
488 & mmlu/high\_school\_world\_history \citep{hendryckstest2021} & MultipleChoice \\
489 & mmlu/college\_physics \citep{hendryckstest2021} & MultipleChoice \\
490 & mmlu/high\_school\_us\_history \citep{hendryckstest2021} & MultipleChoice \\
491 & mmlu/high\_school\_statistics \citep{hendryckstest2021} & MultipleChoice \\
492 & mmlu/conceptual\_physics \citep{hendryckstest2021} & MultipleChoice \\
493 & mmlu/high\_school\_psychology \citep{hendryckstest2021} & MultipleChoice \\
494 & mmlu/high\_school\_physics \citep{hendryckstest2021} & MultipleChoice \\
495 & mmlu/high\_school\_microeconomics \citep{hendryckstest2021} & MultipleChoice \\
496 & mmlu/high\_school\_mathematics \citep{hendryckstest2021} & MultipleChoice \\
497 & mmlu/econometrics \citep{hendryckstest2021} & MultipleChoice \\
498 & mmlu/high\_school\_macroeconomics \citep{hendryckstest2021} & MultipleChoice \\
499 & mmlu/high\_school\_government\_and\_politics \citep{hendryckstest2021} & MultipleChoice \\
500 & mmlu/high\_school\_geography \citep{hendryckstest2021} & MultipleChoice \\
501 & mmlu/high\_school\_european\_history \citep{hendryckstest2021} & MultipleChoice \\
502 & mmlu/high\_school\_computer\_science \citep{hendryckstest2021} & MultipleChoice \\
503 & mmlu/high\_school\_chemistry \citep{hendryckstest2021} & MultipleChoice \\
504 & mmlu/high\_school\_biology \citep{hendryckstest2021} & MultipleChoice \\
505 & mmlu/marketing \citep{hendryckstest2021} & MultipleChoice \\
506 & mmlu/management \citep{hendryckstest2021} & MultipleChoice \\
507 & mmlu/moral\_disputes \citep{hendryckstest2021} & MultipleChoice \\
508 & mmlu/formal\_logic \citep{hendryckstest2021} & MultipleChoice \\
509 & mmlu/security\_studies \citep{hendryckstest2021} & MultipleChoice \\
510 & mmlu/public\_relations \citep{hendryckstest2021} & MultipleChoice \\
511 & winogrande/winogrande\_xl \citep{ai2:winogrande} & MultipleChoice \\
512 & codah/codah \citep{chen2019codah} & MultipleChoice \\
513 & ai2\_arc/ARC-Challenge/challenge \citep{allenai:arc} & MultipleChoice \\
514 & ai2\_arc/ARC-Easy/challenge \citep{allenai:arc} & MultipleChoice \\
515 & definite\_pronoun\_resolution \citep{rahman2012resolving} & MultipleChoice \\
516 & swag/regular \citep{zellers2018swagaf} & MultipleChoice \\
517 & math\_qa & MultipleChoice \\
518 & lex\_glue/case\_hold \citep{Zheng2021} & MultipleChoice \\
519 & commonsense\_qa \citep{talmor-etal-2019-commonsenseqa} & MultipleChoice \\
520 & discosense & MultipleChoice \\
521 & medmcqa \citep{pmlr-v174-pal22a} & MultipleChoice \\
522 & aqua\_rat/tokenized \citep{ACL} & MultipleChoice \\
523 & logiqa \citep{liu2020logiqa} & MultipleChoice \\
524 & cycic\_multiplechoice \citep{Kejriwal2020DoFC} & MultipleChoice \\
525 & arct \citep{Habernal.et.al.2018.NAACL.ARCT} & MultipleChoice \\
526 & onestop\_qa \citep{starc2020} & MultipleChoice \\
527 & moral\_stories/full \citep{Emelin2021MoralSS} & MultipleChoice \\
528 & prost \citep{aroca-ouellette-etal-2021-prost} & MultipleChoice \\
529 & webgpt\_comparisons \citep{nakano2021webgpt} & MultipleChoice \\
530 & synthetic-instruct-gptj-pairwise & MultipleChoice \\
531 & wouldyourather & MultipleChoice \\
532 & summarize\_from\_feedback/comparisons \citep{stienon2020learning} & MultipleChoice \\
533 & SHP \citep{SHP} & MultipleChoice \\
534 & MedQA-USMLE-4-options-hf & MultipleChoice \\
535 & wikimedqa/medwiki \citep{sileo2023generating} & MultipleChoice \\
536 & cicero \citep{ghosal2022cicero} & MultipleChoice \\
537 & mutual \citep{mutual} & MultipleChoice \\
538 & NeQA & MultipleChoice \\
539 & quote-repetition & MultipleChoice \\
540 & redefine-math & MultipleChoice \\
541 & implicatures \citep{george2020conversational} & MultipleChoice \\
542 & race/high \citep{lai2017large} & MultipleChoice \\
543 & race/middle \citep{lai2017large} & MultipleChoice \\
544 & race-c \citep{pmlr-v101-liang19a} & MultipleChoice \\
545 & spartqa-mchoice \citep{mirzaee-etal-2021-spartqa} & MultipleChoice \\
546 & riddle\_sense \citep{lin-etal-2021-riddlesense} & MultipleChoice \\
547 & reclor \citep{yu2020reclor} & MultipleChoice \\
548 & ScienceQA\_text\_only \citep{10.1007/s00799-022-00329-y} & MultipleChoice \\
549 & ekar\_english & MultipleChoice \\
550 & path-naturalness-prediction & MultipleChoice \\
551 & cloth & MultipleChoice \\
552 & dgen & MultipleChoice \\
553 & oasst1\_pairwise\_rlhf\_reward & MultipleChoice \\
554 & conll2003/pos\_tags \citep{tjong-kim-sang-de-meulder-2003-introduction} & TokenClassification \\
555 & conll2003/chunk\_tags \citep{tjong-kim-sang-de-meulder-2003-introduction} & TokenClassification \\
556 & conll2003/ner\_tags \citep{tjong-kim-sang-de-meulder-2003-introduction} & TokenClassification \\
557 & wnut\_17/wnut\_17 \citep{derczynski-etal-2017-results} & TokenClassification \\
558 & ncbi\_disease/ncbi\_disease \citep{dougan2014ncbi} & TokenClassification \\
559 & acronym\_identification \citep{veyseh-et-al-2020-what} & TokenClassification \\
560 & jnlpba/jnlpba \citep{kim2004introduction} & TokenClassification \\
561 & species\_800/species\_800 \citep{pafilis2013species} & TokenClassification \\
562 & ontonotes\_english \citep{tjong-kim-sang-de-meulder-2003-introduction} & TokenClassification \\
563 & universal\_dependencies/en\_partut/deprel \citep{11234/1-3424} & TokenClassification \\
564 & universal\_dependencies/en\_lines/deprel \citep{11234/1-3424} & TokenClassification \\
565 & universal\_dependencies/en\_gumreddit/deprel \citep{11234/1-3424} & TokenClassification \\
566 & universal\_dependencies/en\_esl/deprel \citep{11234/1-3424} & TokenClassification \\
567 & universal\_dependencies/en\_ewt/deprel \citep{11234/1-3424} & TokenClassification \\
568 & universal\_dependencies/en\_gum/deprel \citep{11234/1-3424} & TokenClassification \\
\end{longtable}
\nocite{sileo-lmrec-2022, laurer2022less}

\newpage
\section{Model Recycling results}
\begin{table}[H]
\centering
\begin{tabular}{@{}lll@{}}
\toprule
model\_name            & deberta-v3-base & +tasksource \\ \midrule
avg                    & 79.04           & \textbf{80.73}             \\
mnli (linear probe)               & -               & \textbf{93.73}             \\
20\_newsgroup          & 86.41           & \textbf{86.46}             \\
ag\_news               & 90.44           & \textbf{90.67}             \\
amazon\_reviews\_multi & 66.86           & \textbf{66.90}             \\
anli                   & 58.78           & \textbf{60.38}             \\
boolq                  & 82.99           & \textbf{85.66}             \\
cb                     & 75.00           & \textbf{82.14}             \\
cola                   & 86.57           & \textbf{87.15}             \\
copa                   & 58.40           & \textbf{81.00}             \\
dbpedia                & \textbf{79.43}  & 79.20                      \\
esnli                  & 91.93           & \textbf{91.54}             \\
financial\_phrasebank  & 84.48           & \textbf{85.20}             \\
imdb                   & 94.49           & \textbf{94.67}             \\
isear                  & 71.86           & \textbf{71.90}             \\
mnli\_mismatched       & 89.78           & \textbf{91.14}             \\
mrpc                   & \textbf{89.20}  & 88.73                      \\
multirc                & 62.26           & \textbf{63.82}             \\
poem\_sentiment        & 86.73           & \textbf{92.31}             \\
qnli                   & 93.51           & \textbf{93.72}             \\
qqp                    & 91.79           & \textbf{91.92}             \\
rotten\_tomatoes       & 90.42           & \textbf{90.99}             \\
rte                    & 82.35           & \textbf{90.61}             \\
sst2                   & 95.06           & \textbf{95.41}             \\
sst\_5bins             & 56.98           & \textbf{58.60}             \\
stsb                   & 90.28           & \textbf{91.81}             \\
trec\_coarse           & 97.76           & \textbf{96.80}             \\
trec\_fine             & \textbf{91.02}  & 90.80                      \\
tweet\_ev\_emoji       & 46.19           & \textbf{47.82}             \\
tweet\_ev\_emotion     & 83.95           & \textbf{85.71}             \\
tweet\_ev\_hate        & 56.21           & \textbf{57.47}             \\
tweet\_ev\_irony       & 79.82           & \textbf{83.04}             \\
tweet\_ev\_offensive   & 85.06           & \textbf{85.23}             \\
tweet\_ev\_sentiment   & 71.80           & \textbf{72.01}             \\
wic                    & \textbf{71.21}  & 69.44                      \\
wnli                   & \textbf{70.21}  & 67.61                      \\
wsc                    & 64.09           & \textbf{66.35}             \\
yahoo\_answers         & 72.03           & \textbf{72.07}             \\ \bottomrule
\end{tabular}
\end{table}
\end{document}